# NUVA: A Naming Utterance Verifier for Aphasia Treatment

*David S. Barbera[1], Mark Huckvale[2], Victoria Fleming[1], Emily Upton[1], Henry Coley-Fisher[1], Catherine Doogan[1], Ian Shaw[3], William Latham[4], Alexander P. Leff[1], Jenny Crinion[1]*

[1]Institute of Cognitive Neuroscience, University College London, U.K.
[2]Speech, Hearing & Phonetic Sciences, University College London, U.K.
[3]Technical Consultant at SoftV, U.K.
[4]Goldsmiths College, University of London, U.K.
david.barbera.16@ucl.ac.uk

## Abstract:

Anomia (word-finding difficulties) is the hallmark of aphasia, an acquired language disorder most commonly caused by stroke. Assessment of speech performance using picture naming tasks is a key method for both diagnosis and monitoring of responses to treatment interventions by people with aphasia (PWA). Currently, this assessment is conducted manually by speech and language therapists (SLT). Surprisingly, despite advancements in automatic speech recognition (ASR) and artificial intelligence with technologies like deep learning, research on developing automated systems for this task has been scarce. Here we present NUVA, an utterance verification system incorporating a deep learning element that classifies 'correct' versus' incorrect' naming attempts from aphasic stroke patients. When tested on eight native British-English speaking PWA the system's performance accuracy ranged between 83.6% to 93.6%, with a 10-fold cross-validation mean of 89.5%. This performance was not only significantly better than a baseline created for this study using one of the leading commercially available ASRs (Google speech-to-text service) but also comparable in some instances with two independent SLT ratings for the same dataset.

**Index Terms**: speech disorders, word naming, aphasia, anomia, speech recognition, Dynamic Time Warping

## 1. Introduction

Word retrieval difficulties(aka anomia) are one of the most pervasive symptoms of post-stroke aphasia (Matti. Laine, 2006). Recent data suggests there are around 350,000 people in the UK alone living with chronic aphasia (*Stroke Association*, 2018). Despite its' prevalence, few individuals receive a sufficient dose of speech and language therapy to recover maximally. For example, in the UK through the National Health Service, patients receive on average 8-12 hours (Code & Heron, 2003); however, reviews of speech and language intervention studies have shown superior outcomes for treatments that deliver around 100 hours of therapy (Bhogal et al., 2003; Brady et al., 2016). Assessment of patients' spoken picture naming abilities and then practising repetitively over time a range of vocabulary using spoken picture naming tasks, are both integral parts of impairment based anomia treatments (Whitworth et al., 2014).

Picture naming tasks (confrontation naming) are typically carried out in face-to-face settings, to allow SLTs to assess the individual's accuracy for each naming attempt, and, crucially, provide item-by-item feedback during therapy sessions. An Automated Speech Recognition system (ASR) that could reliably assess patients' speech performance on these picture naming tasks would offer two clear advantages. Firstly, increased consistency across naming attempts, and sensitivity to changes in patients' speech abilities over time, would enable therapeutic interventions can be better tracked. Secondly, ASR would allow patients to be able to perform these tasks independent of SLTs potentially remotely away from the clinic in the comfort of their own home. This would not only 'free-up' SLTs to deliver more complex interventions in their 'face-to-face' clinical time but also support more patients who are unable to travel into the clinic to use these tasks effectively, a need which has become more pressing in light of recent COVID-19 travel restrictions and social distancing.

## 1.1 ASR in disordered speech

To date, most ASRs applied to speech disorders have targeted people with dysarthria - a motor speech disorder. While this can also be caused by stroke and patients may manifest both dysarthria and aphasia concurrently, the nature and characteristics of the speech deficits in dysarthria and aphasia are quite different (Abad et al., 2013). Dysarthria is a speech disorder resulting from a weakness, paralysis, or incoordination of the speech musculature (Darley Frederic L. et al., 1969). In this disorder, patients know the target word, e.g., spoon and correctly retrieve the lexical item, but the sounds are distorted, e.g., /poon/. These articulation errors tend to be highly consistent in a pattern, e.g., a patient may have difficulties saying all words beginning with an /s/, irrespective of the type of word, due to impaired tongue movements. In contrast, aphasia is an acquired disorder of language that is caused by damage to certain areas of the brain, which are primarily responsible for language function (Akbarzadeh-T & Moshtagh-Khorasani, 2007). Patients with this disorder commonly report that they know what they want to say but just can't find the right words. They do not have a motor speech disorder (i.e. they can say the words and sounds correctly, repetition can be intact), but rather, have a word (lexical) retrieval problem, akin to the 'tip of the tongue' phenomenon experienced by non-aphasic individuals. Alongside this, patients can make a variety of errors when speaking; for example, they may want to say /tea/ but say /coffee/ instead, a semantically related word, or say /key/, a phonemically related word, or even say a non-word, e.g., /tife/. There is high variability in speech error types, both between aphasic individuals with the same disorder and within an individual over time, even within the same day. This variety in speech patterns is a challenge for ASR systems and perhaps explains why the use and study of ASR in aphasia has been surprisingly scarce to date.

## 1.2 Overview of ASR in aphasia

The earliest studies to explore ASR in aphasia adopted 'off-the-shelf' commercially available software developed for healthy speakers. They utilised these tools to provide automated speech performance feedback to aphasic patients during therapy with mixed results. Linebarger and colleagues (Linebarger et al., 2001) incorporated a speech recognition module as part of a computerised-based training for language production. They reported positive treatment outcomes for the patients but unfortunately not much detail about the utility of the ASR technology used. Wade and colleagues (J. Wade, 2001) investigated the use of speech recognition software in aphasia and reported ASR accuracy levels of 80% at single word level and at phrase level. The software required extensive training sessions to adapt to the aphasic speaker and utilised a limited vocabulary of 50 words and 24 sentences. Further improvements in the reliability and stability of the system were recommended before clinical implementation.

Having identified the limitations of the previous approaches and the large unmet clinical need, more recent research efforts have focused on developing new ASR technology that can be used more effectively in aphasia. The latest advances in the ASR field and a prominent use of deep neural networks (DNN) have enabled a small number of groups to achieve: (1) transcriptions of narrative speech by PWA in both English and Cantonese (Le & Provost, 2016; T. Lee et al., 2016); (2) preliminary detection of phonemic paraphasic speech errors in connected speech by PWA (Le et al., 2017); (3) quantitative analysis of spontaneous speech by PWA (Le et al., 2018); and, (4) assessment of speech impairment in Cantonese PWA (Qin et al., 2019). Nevertheless, these current ASR solutions remain susceptible to variations in the language spoken, e.g., dialects and accents. While large corpora of aphasic speech samples in openly available databases, e.g., AphasiaBank in USA (Forbes et al., 2012;

MacWhinney et al., 2011) are being developed that will provide useful priors/training sets for new models in the future, it is likely that not all languages or accents will benefit equally from this approach.

To address the lack of globally relevant aphasic speech corpora, and to develop a system that is able to take account of the high variability of speech performance by PWA and the distinction between aphasic speech and dysarthric speech we aimed to develop an ASR system that could reliably assess spoken single word picture naming performance in patients with aphasia without motor speech impairment (dysarthria or apraxia).

### 1.3 ASR for single word naming performance by PWA

In contrast to the challenge of analysing spontaneous speech utterances, assessing spoken picture naming performance has the advantage that the target word is known. The challenge for ASR in this context is to verify that a certain target word is uttered in a given segment of speech (Abad et al., 2013). Within the context of an intervention, the ASR system, or utterance verifier system, must also process each utterance in quick succession so that a binary 'correct'/'incorrect' response can provide feedback to the PWA or the therapy governing algorithm on an item-by-item basis.
To the best of our knowledge, only two groups have used and assessed an ASR-based system of such type in aphasic speakers' single word picture naming performance. In the project Vithea (Pompili et al., 2011), researchers developed an app for the treatment of aphasia for Portuguese speakers. They first presented results of an in-house ASR-engine called AUDIMUS (Meinedo et al., 2003) assessing picture naming using a keyword spotting technique to score spoken naming attempts as 'correct'/'incorrect'. An updated version of this reported an average ASR accuracy of 82%, with ranges between 69% and 93% across patients (Abad et al., 2013). The second group (Ballard et al., 2019) evaluated a digitally delivered intervention using picture naming tasks in native Australian English speaking people with both apraxia and aphasia. They used the open-source ASR engine CMU PocketSphinx (*Cmusphinx/Pocketsphinx*, 2014/2020) to provide patients with 'correct'/'incorrect' feedback for each of their naming attempts during treatment. The ASR-engine recognized 124 words; each word was phonetically different from the others. They reported an overall system performance accuracy of 80% and a range of well-classified scores between 65.1% and 82.8% across patients depending on impairment severity. Both these systems provided good 'proof-of-concept 'data that ASR systems for assessing spoken word performance are feasible, but the high error rate and variable performance across aphasic patients meant their clinical utility remained low.

The aim of this project is to present and assess the feasibility and stability of NUVA, a tailor-made ASR system incorporating a deep learning element to assess word naming attempts in people with aphasia.

## 2. Method

### 2.1 Deep Learning and Recurrent Neural Networks

As a subfield of artificial intelligence, machine learning studies computer algorithms that are able to learn from data how to perform a certain task (Goodfellow et al., 2016). Deep learning is a specific kind of machine learning which allows computational models that are composed of multiple layers to learn representations of data with multiple levels of abstraction, and has dramatically improved the state-of-the-art in speech recognition, visual object recognition, and many other domains such as drug discovery and genomics (LeCun et al., 2015). In its supervised form, these models are trained using data that comes with the ground truth -or labels- of the specific task the algorithm aims to learn. For example, in a phone-recognition task, the algorithm learns from speech signals (the data) where phoneticians have placed manually the phone boundaries as time alignments (the labels). Deep learning models are also called deep neural networks (DNN) as their layers take the form of artificial neural networks. For tasks that involve sequential data, such as speech, it is often better to use recurrent neural networks (RNN): at every time-step, each element of a sequence is processed by a RNN keeping track in their hidden units a 'state vector' that implicitly contains information about the history of all past elements of the sequence (LeCun et al., 2015). RNNs can be configured to take into consideration all the future elements of a sequence in addition to the past ones, in which case they are called bidirectional RNNs. However, the recurrent nature of this type of neural network can cause its hidden units to 'blow up' or vanish due to error signals flowing backward in time. To overcome such problem, a variation of RNNs was introduced called long short-term memory (LSTM) in which memory cells and gate units are used to construct an architecture that allows for constant error flow (Hochreiter & Schmidhuber, 1997). More recently, a variation of LSTMs was proposed using gate units without memory cells, called gated recurrent units (GRUs) showing comparable performance on speech signal modelling (Cho et al., 2014; Chung et al., 2014).

## 2.2 NUVA: A Naming Utterance Verifier for Aphasia

Given the scarcity of speech corpora in aphasia, we used a template-based system for picture naming verification. Key to our approach was utilising the framework developed by Ann Lee and James Glass (A. Lee & Glass, 2012) to detect word-level mispronunciations in non-native speech. It uses posteriorgram based pattern matching via a dynamic time warping (DTW) algorithm to compare a word uttered by a native speaker, (teacher), with the same word uttered by a non-native speaker (student). It was designed to be language agnostic. To generate posteriorgrams, our system NUVA replaces their Gaussian Mixture Model trained on unlabelled speech corpora with an acoustic model to yield phone-based posteriorgrams using a deep neural network (DNN) trained on English corpora from healthy speakers. Then, similar to Lee's teacher-versus-student framework, we compare healthy-versus-aphasic utterances. We defined a posteriorgram as a vector of posterior probabilities over phoneme classes in the English language for which we employed the ARPAbet system as used in the BEEP dictionary (Robinson, 1996) consisting of 45 symbols: 44 ARPAbet symbols plus silence. To enable future clinical utility of NUVA, we developed it to run embedded on mobile devices without computationally sophisticated model compression techniques.

DNNs together with DTW has been recently used to solve the task of query-by-example, where a system aims to find the closest matching phrase (Anguera et al., 2015; Ram et al., 2020; Yuan et al., 2020). However, in our task a system just needs to verify that a phrase is what it claims to be.

The following sections describe our system, how speech is processed, how our models were trained, and how our 'utterance-verifier' system NUVA was assembled. We then report the performance of NUVA at classifying eight aphasic patients' spoken picture naming attempts as 'correct' or 'incorrect' compared to SLT classifications of the same responses. SLT classification is seen as the clinical 'gold-standard' approach, and our aim was to develop an automated utterance verification system whose performance could match the 'gold-standard'.

## 2.3 Signal pre-processing and acoustic modelling

Speech recordings were pre-processed in overlapping frames of 30 milliseconds every 10 milliseconds, and a fast Fourier transform size of 64 milliseconds after a pre-emphasis filter of 0.95 to obtain a vector of 26 acoustic features per frame: 12 Mel-frequency cepstral coefficients (with a final liftering step with a coefficient of 23 applied to them), energy and 13 deltas. See step 1 and 2 in Figure 1.

To train our acoustic model, we used a corpus of British healthy English speakers WSJCAM0 (Robinson et al., 1995). WSJCAM offers phone-level transcriptions using the ARPAbet phone set for British English. We then used Keras deep learning framework (Chollet & others, 2015) with TensorFlow (Martín Abadi et al., 2015) as the back-end. All our models used batch normalisation, a dropout rate of 0.5 and a categorical cross-entropy over 45 classes as the loss function. The training lasted until there was no improvement in accuracy for 50 epochs. We explored several types and configurations of recurrent neural networks and chose our final model as the one with the lowest Phone Error Rate (PER) on the WSJCAM0 test set. Our winning model was a bidirectional Gated Recurrent Unit (Chung et al., 2014) of 128 units and seven layers of depth trained with the Adam optimizer (Kingma & Ba, 2014) resulting in around 2 million parameters and achieving a phone error rate (PER) of 15.85%. See step 3 in Figure 1.

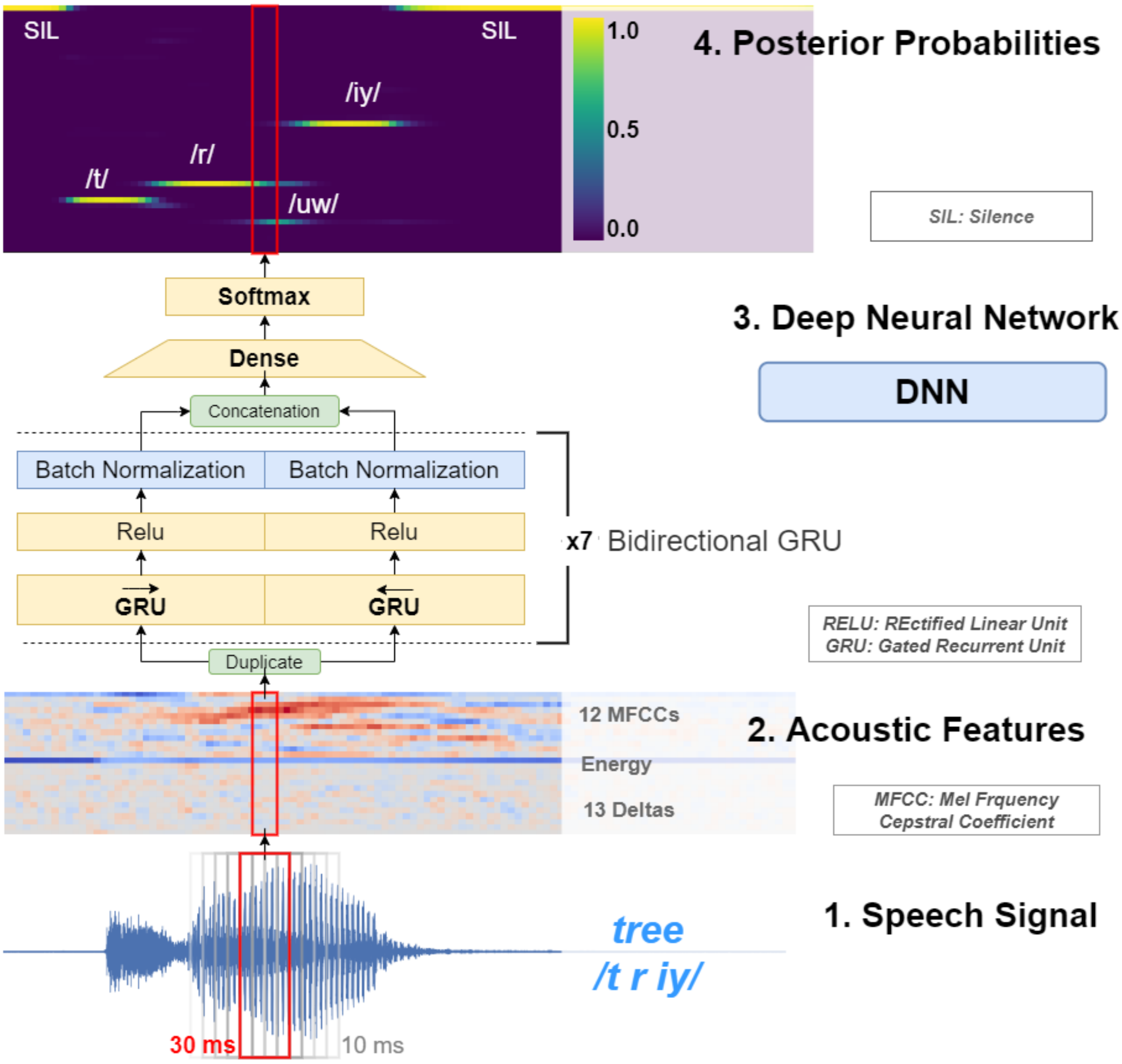

*Figure 1. From signal to posterior probabilities. Bottom to top: speech signal is fragmented into frames every 10 milliseconds of a window size of 30 milliseconds (1), from each frame a vector of acoustic features is extracted (2) then each vector is fed to a Deep Neural Network (3) which outputs a vector of posterior probabilities or posteriorgram (4).*

## 2.4 Comparison of utterances

NUVA uses two recordings from healthy native speakers for each target word, which are transformed into posteriorgrams offline via our DNN, as shown in Figure 1 (steps 1-4). Each naming attempt by an aphasic speaker is transformed into posteriorgrams online using our DNN and then compared to each of the posteriorgrams from the two healthy speakers via the DTW algorithm, see Figure 2. Adapting Lee's notation, (Ann Lee & Glass, 2012) given a sequence of posteriorgrams for the healthy speaker $H = (p_{h_1}, p_{h_2}, \dots, p_{h_n})$ and the aphasic speaker $A = (p_{a_1}, p_{a_2}, \dots, p_{a_m})$, a $n \times m$ distance matrix can be defined using the following inner product:

$$\varphi_{ha}(i,j) = -\log(p_{h_i} * p_{a_j}) \quad (1)$$

For such a distance matrix, DTW will search for the path from $(1,1)$ to $(n,m)$ that minimizes the accumulated distance. Different from Lee's work, we used the minimum of the DTW accumulated distances for all comparisons with the two healthy speakers to make a final decision.

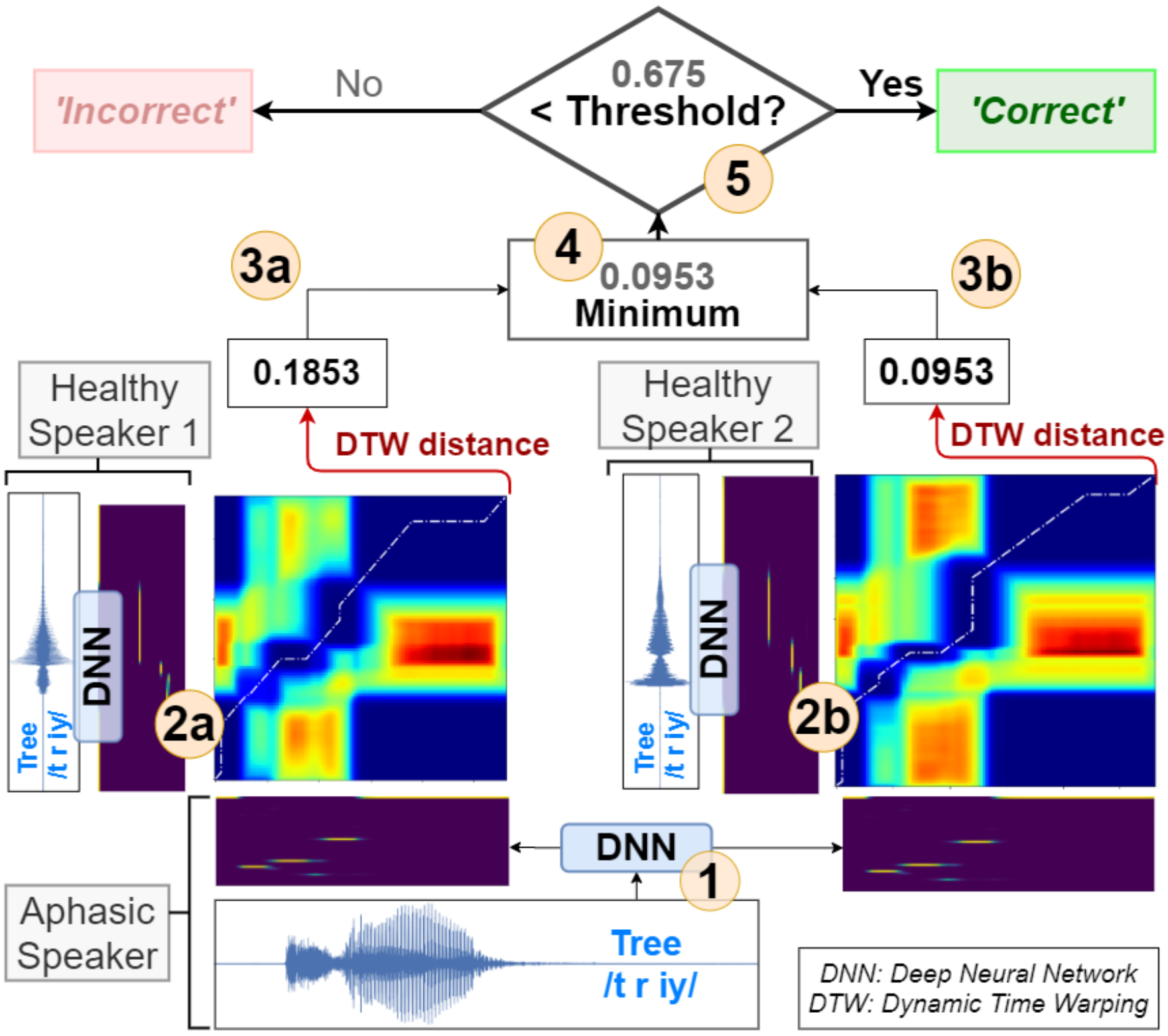

*Figure 2. NUVA: An utterance verification system for assessing naming attempts in anomia treatment. Given a naming attempt, e.g. target word /tree/, the aphasic's utterance is recorded and processed through our DNN) to generate posteriorgrams (1). The system keeps posteriorgrams of previously recorded healthy speakers' utterances for each target word, (2a and 2b). Posteriorgrams are compared using the DTW) algorithm yielding a distance number between 0 and +∞ (3a and 3b). The minimum of both distances is selected (4) and compared to a set threshold (5) calibrated per speaker, in this example 0.675. If the distance is less than the threshold, then the decision is that the aphasic speaker has said the target word correctly, otherwise it is classified as incorrect.*

Many biometric systems of verification utilise a threshold value to make a decision (Malik et al., 2014). In NUVA if the value is smaller than the set threshold then the aphasic's naming attempt is deemed similar enough to that of the healthy speakers and is considered 'correct'; otherwise, it is classified as 'incorrect', see Figure 2. The existence of such threshold is functionally analogous to the $\beta$ penalizing parameter in Abad (Abad et al., 2013): calibration of both controls the rate of false positives and false negatives for each patient. However, they are intrinsically different within each system.

The reason for using the minimum of DTW distances between the healthy speakers can intuitively be justified as choosing the healthy speaker's utterance that is closest aligned to the aphasic speaker. The number of healthy speakers could be increased to cater for a broader range of accents, offering a closer regional match to each aphasic speaker. Figure 3 illustrates this rationale quantitatively using Receiver Operating Characteristic (ROC) curves and Area Under the Roc Curve (AUC) scores. Choosing the minimum of the DTW distances between the healthy speakers yields the best fit and highest AUC score to the scorings provided by a speech and language therapist (SLT) used in this study. We discuss this further in the next section.

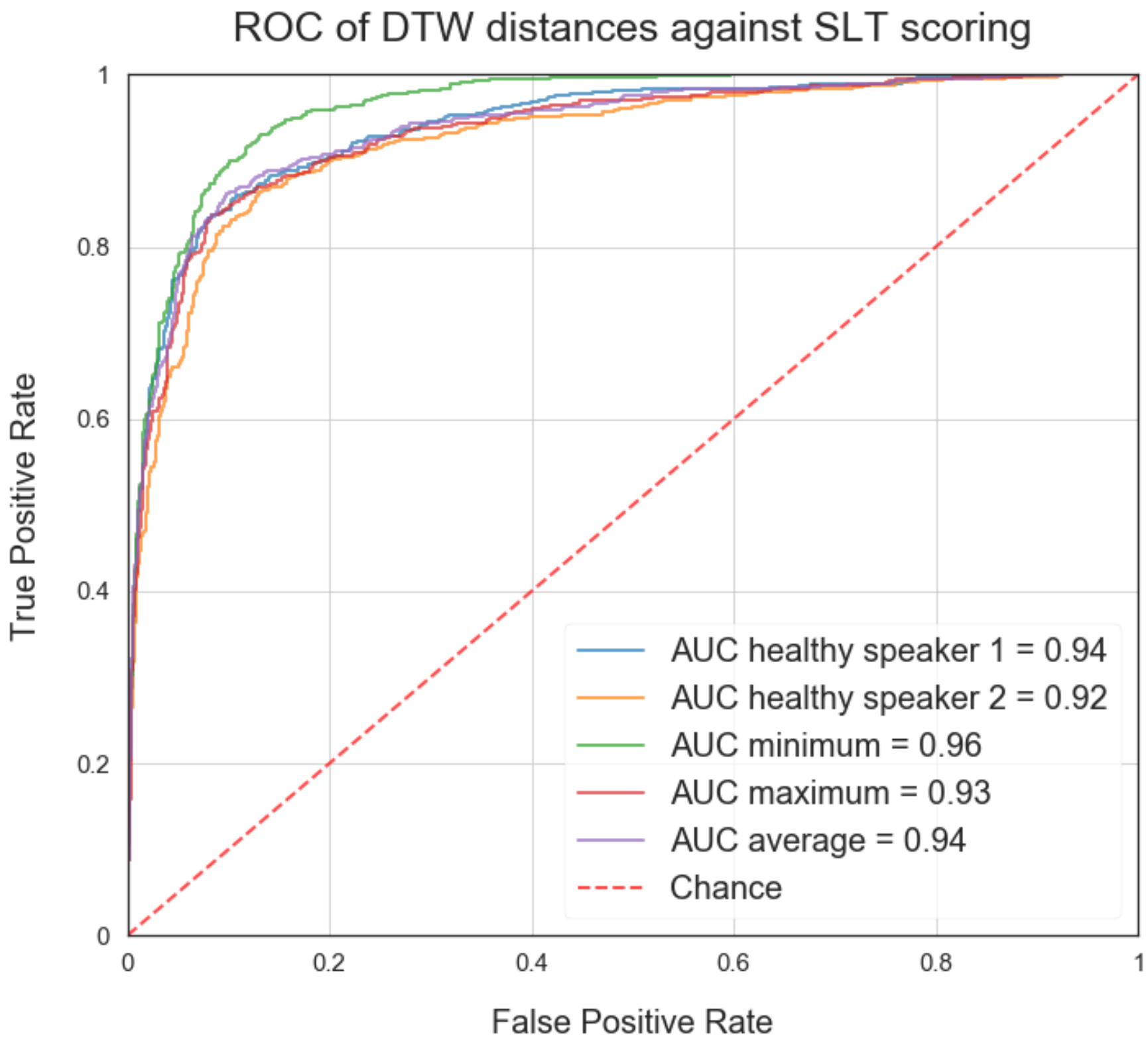

*Figure 3. ROC curves for SLT scoring. Each curve represents the best fit to SLT scorings for a given criteria of combining DTW distances: (i) choosing one healthy speaker, (ii) the minimum distance between healthy speakers, (iii) the maximum and (iv) the average.*

## 2.5 Threshold Calibration

NUVA requires a set threshold to make a decision to determine whether an attempt is 'correct' or 'incorrect'. Previous naming data from a PWA as assessed by their clinical SLT could be used for such purposes. In real life, a PWA is first assessed by a SLT to determine their diagnosis and level of speech and language impairment before being referred to a specific treatment program. Speech performance on picture naming tasks is a core part of the standard clinical SLT aphasia assessment, so it is reasonable to assume the existence of previously labelled (by a SLT) speech data may be available to calibrate NUVA's threshold. Two methods used to calibrate the most adequate threshold given previously held data are:

1. **Adapted per patient**

The threshold is calibrated using previously held data from a certain patient. In this case, every new patient will need to have the threshold calibrated according to the data used in a previous assessment.

2. **Fixed**

The threshold is calibrated using previously held data from various patients and then is fixed for any new patient.

## 3. Experiment and data

### 3.1 Healthy speakers corpus: WSJCAM0

The underlying DNN of our system was trained using speech from healthy English speakers. We used WSJCAM0, a corpus of spoken British English derived from the Wall Street Journal text corpus specifically designed for the construction and evaluation of speaker-independent speech recognition systems (Robinson et al., 1995). The database has recordings of 140 speakers each speaking about 110 utterances. The corpus is partitioned into 92 training speakers, 20 development test speakers and two sets of 14 evaluation test speakers. It offers word and phone based time alignments on verified utterance transcriptions and employs ARPAbet as a phone set. WSJCAM0 is currently distributed by the Linguistic Data Consortium.

### 3.2 Participants

Eight native English speakers, six male, with chronic post-stroke anomia, were recruited. Demographics are shown in Table 1 below. Inclusion criteria were aphasia in the absence of speech apraxia (severe motor speech impairment) as evidenced by: (i) impaired naming ability on the object naming subtest of the Comprehensive Aphasia Test, CAT (Swinburn et al, 2004); scores below < 38 are classified as impaired; (ii) good single word repetition on the CAT subtest; normative cut-off>12. All patients gave written consent, and data was processed in accordance with current GDPR guidelines. Ethical approval was granted by NRES Committee East of England – Cambridge, 18/EE/228.

Table 1. Demographic and clinical data for the patient.

| Patient ID | Sex | Age | Months post-stroke | CAT Object naming | CAT Repetition |
|---|---|---|---|---|---|
| **P1** | M | 65 | 108 | 32 | 19 |
| **P2** | M | 58 | 90 | 19 | 22 |
| **P3** | M | 70 | 91 | 10 | 28 |
| **P4** | F | 62 | 21 | 28 | 24 |
| **P5** | M | 64 | 14 | 6 | 25 |
| **P6** | M | 59 | 98 | 30 | 31 |
| **P7** | M | 57 | 109 | 27 | 24 |
| **P8** | F | 82 | 38 | 29 | 23 |
| **Mean(IQR)** | | 65(8) | 71(67) | 23(13) | 25(3) |
| | | | **Max score possible** | (/48) | (/32) |
| | | | **Cut-off used** | <38 | >12 |

### 3.3 Stimuli

Picture naming stimuli consisted of 220 coloured drawings. They were selected from the top 2000 most frequent words using the *Zipf* index of the SUBTLEX-UK corpus (van Heuven et al., 2014) keeping the same distribution of parts of speech for nouns, verbs, adverbs and adjectives. See Appendix C for a list of the items selected.

### 3.4 Dataset Collection

We used a tailor-made gamified picture naming treatment App developed in Unity on an Android tablet Samsung SM-T820 to deliver the picture stimuli and record the patients' speech responses. Patients' speech recordings were collected using a Sennheiser headset SC 665 USB at 16 kHz which were then stored in a compliant WAVE-formatted file using a 16 bit PCM encoding.

Patients were instructed to name each item presented on screen as quickly and accurately as possible using a single word response. They were given up to 6 seconds to complete each picture naming attempt. The item presentation order was randomised across patients. A SLT was present throughout the assessment and scored the naming responses online in a separate file without giving the patient any performance feedback. A total of 1760 speech recordings (220 words x 8 patients) were acquired.

## 3.5 Procedure

A SLT classified all naming attempts online while patients were in session with the treatment App. After a further revision, six categories were used to classify the attempts using the transcriptions and notes taken during the online session: "Correct", "No Response", "Filler", "Phonological Error", "Circumlocution" and "Other", in addition to “Correct” or “Incorrect” as a final decision. These six categories were used a posteriori to manually detect speech sections within each recording that were used by the SLT to make a decision. When patients produced multiple utterances in a naming attempt, only one utterance representative of the online verdict by the SLT of such naming attempt was segmented to construct a single-utterance recording. For example, when a patient's response was scored as 'Filler', and the corresponding recording comprised of multiple 'um', 'ah', 'eh', only one of those utterances was selected and segmented to create a single-utterance naming attempt per item. These single-utterance recordings were the data used to evaluate NUVA and the baseline. Each naming attempt was then relabelled as 'correct' or 'incorrect' accordingly, and this last classification was used as the ground truth to evaluate NUVA's performance and baseline. Figure 4 describes the dataset and each of the patient's naming performance. Performances were calculated using 'Scikit-learn' (Pedregosa et al., 2011) and significance testing was implemented using 'MLxtend' (Raschka, 2018) and 'scikit-posthocs' (Terpilowski, 2019) packages.

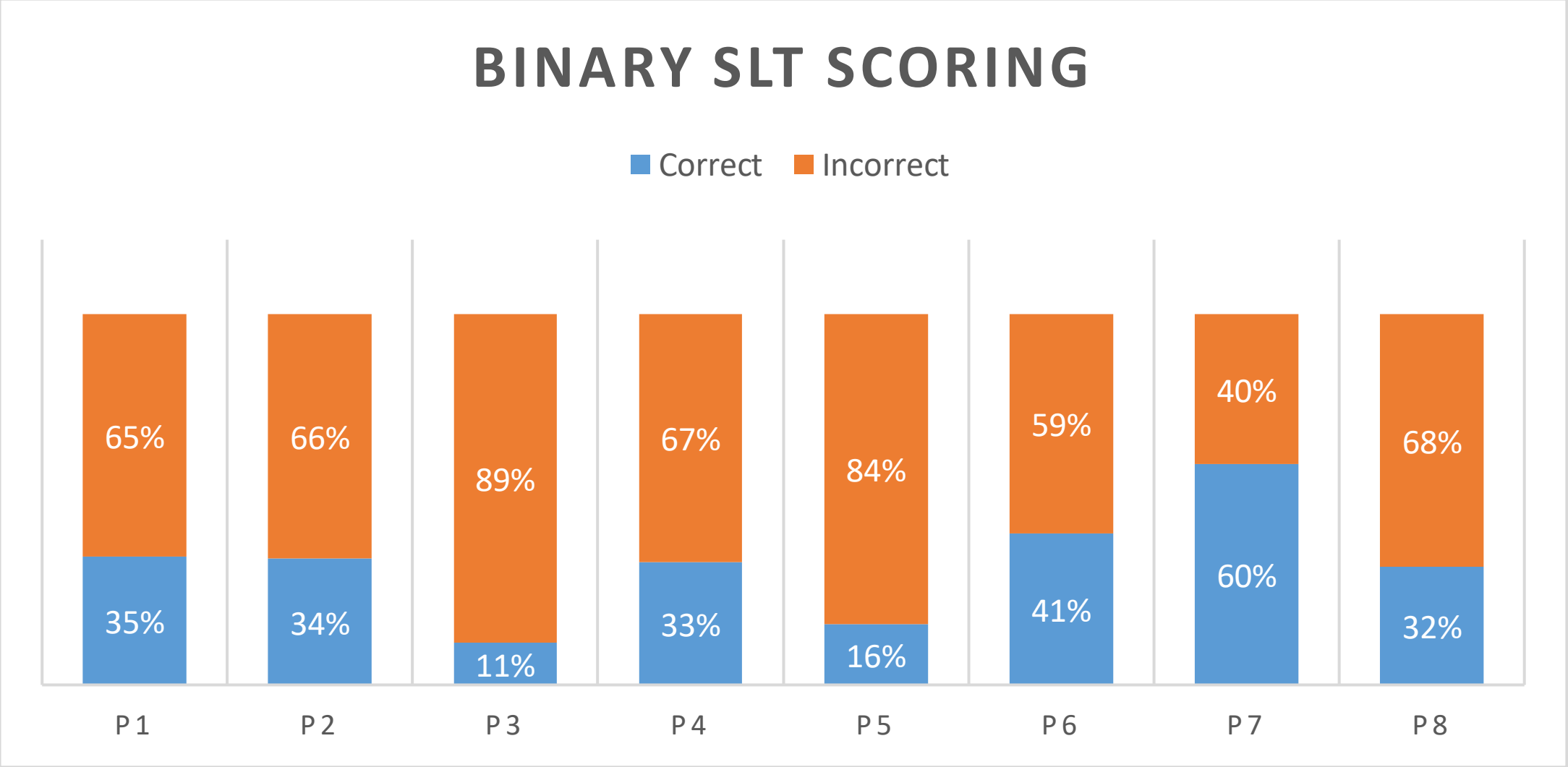

*Figure 4. Each patient's naming performance on the 220 items of the picture naming test, as classified by a speech and language therapist (SLT).*

### *3.5.1 Inter-rater reliability and agreement*

A ground truth is always required to evaluate the performance of an automatic system. In this study, the ground truth was defined as a native English, speaking speech and language therapist’s (SLT) online scoring of the PWAs’ naming attempts. The goal being for the automatic system to be as good as the SLT expert. However, as with all human systems scores might not be identical across individual SLTs. For example, SLTs might vary in determining whether derivatives of the target word, such as singularisation, pluralisation, nominalisation or verbalisation, are classified as correct responses. For this reason, a second native English speaking SLT (SLT2), independently scored all patients' naming attempts offline in addition to the primary SLT who scored them online. In this case, agreement overall between SLT experts is expected to be high, with different scores occurring only on a small number of responses reflecting variance in scoring criteria rather than errors per se. To obtain a measure of inter-rater reliability and account for such measurement noise we used Gwet's AC1 agreement coefficient (Gwet, 2014) i.e., an agreement coefficient corrected by chance.

We then combined this with McHugh's benchmark range for a more stringent interpretation of agreement coefficients as recommended for applications in health care and clinical research (McHugh, 2012) i.e., <.20: No agreement; .21-.39: Minimal; .40-.59: Weak; .60-.79: Moderate; .80-.90: Strong; Above .90: Almost perfect agreement. This was calculated in a series of steps detailed in the Appendix B.2. First, we calculated the probability that the statistic AC1 falls into each of the intervals of the benchmark levels resulting in a membership probability for each interval. Second, we computed the cumulative probability for each interval starting from the highest level of agreement down to the lowest. Third, we took the interval associated with the smallest cumulative probability that exceeds 95% in descending order of agreement as the final benchmark level. The advantage of using this method is that it gives us 95% certainty of the level of agreement reached among the SLT raters. Final level of inter-rater reliability and agreement between both SLTs is summarised in Table 2.

As predicted Inter-rater reliability between SLTs was high for both Gwet's AC1 (mean: 0.938; range 0.85-0.992) and percentage agreement (mean: 96.5%; range 92.3-99.5), with strong to almost perfect agreement.

*Table 2. Inter-rater reliability information for the two SLTs engaged in this study. Showing the Gwet's AC1 agreement coefficient statistic, its 95% confidence interval, level of agreement reached and percentage agreement between the SLTs.*

| Patient | PA[a] | Gwet's AC1 | StdErr[b] | 95% CI[c] | Level of Agreement |
|---|---|---|---|---|---|
| P1 | 99.55% | 0.992 | 0.008 | (0.976, 1.008) | Almost Perfect |
| P2 | 95.91% | 0.925 | 0.025 | (0.877, 0.973) | Strong |
| P3 | 97.73% | 0.972 | 0.012 | (0.948, 0.996) | Almost Perfect |
| P4 | 97.27% | 0.950 | 0.020 | (0.911, 0.990) | Almost Perfect |
| P5 | 97.27% | 0.963 | 0.015 | (0.933, 0.992) | Almost Perfect |
| P6 | 95.45% | 0.913 | 0.027 | (0.860, 0.966) | Strong |
| P7 | 92.27% | 0.850 | 0.035 | (0.781, 0.918) | Moderate |
| P8 | 96.82% | 0.943 | 0.021 | (0.902, 0.985) | Almost Perfect |
| All | 96.53% | 0.938 | 0.008 | (0.923, 0.953) | Almost Perfect |

*[a]PA = percentage agreement*
*[b]StdErr = standard error*
*[c]CI = confidence interval*

Since both, percentage agreement and Gwet's AC1 are very high, we report results using the primary SLT who scored the attempts online as the 'gold-standard' and ground truth for all subsequent analyses with subsequent offline systems. Performance of all offline systems, i.e., the second SLT (SLT2), NUVA and the baseline ASR system were calculated across all reported metrics (Accuracy, False Positives, False Negatives, F1-score and Pearson's r).

#### 3.5.2 ASR Baseline

Since there are no existing baselines that we could use to compare the performance of our system, we used an off-the-shelf standard speech recognition service. The rationale in doing so, is that it would be the fastest way for a clinical lab to implement automatic digital therapy for word naming.

We used Google Cloud Platform speech-to-text service configured with British English (the date used: 24/3/20), and adapted it for our task to create a baseline. As the service provides transcriptions, if the target word was found in the transcript, then the attempt was classified as 'correct', otherwise 'incorrect'. For each aphasic patient's naming attempt, the same recording to test our system was used to test the baseline.

# 4. Results

## 4.1 System Performance

As indicated in section 2.5 Threshold Calibration, NUVA utilises a set threshold to make a final decision on marking a patient's naming attempt as either 'correct' or 'incorrect' and, such set threshold can be calibrated using two methods: 'fixed' and 'adapted', using previously held data. In our study, we used all the available data as held data per each calibration method to have a notion of the best possible performance achievable by NUVA in ideal conditions. With the 'fixed' method, we set the same threshold for all patients to estimate the best possible performance across all patients involved in the study. With the 'adapted' method we set a different threshold for each patient to estimate the best possible performance for each patient. Figure 5 illustrates the performance for all patients across a continuum set of thresholds using Pearson's r as a metric. Figure 6 shows the same performance measure across the same continuum set of thresholds but for each patient. The resulting best thresholds are summarised in Table 3. Thresholds used for each version of NUVA (i) 'fixed' and (ii) 'adapted'.

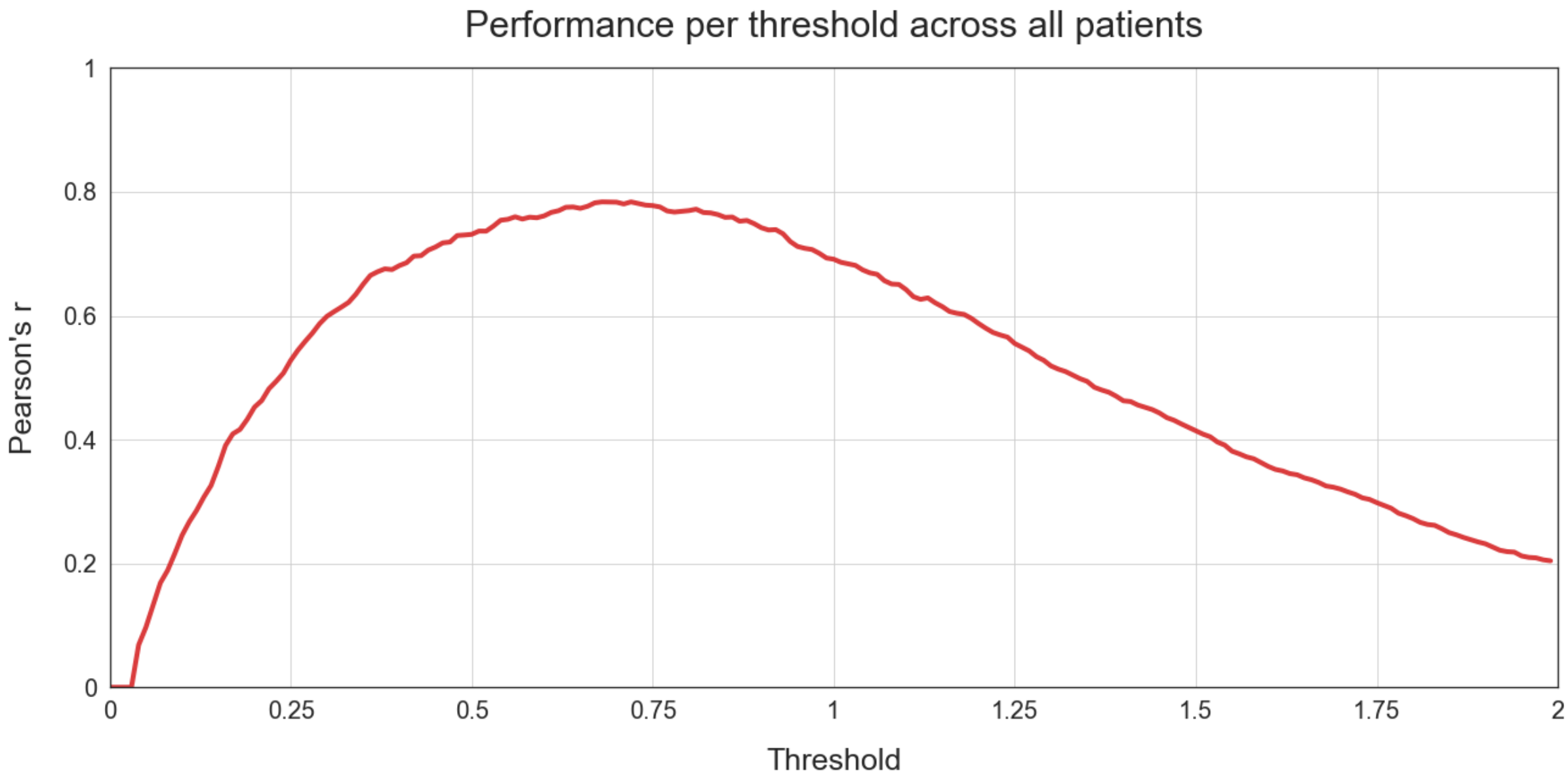

*Figure 5. Performance (y-axis) per threshold (x-axis) across all patients (as a group) using Pearson's r as a metric. Y-axis = closer to 1, the better the performance.*

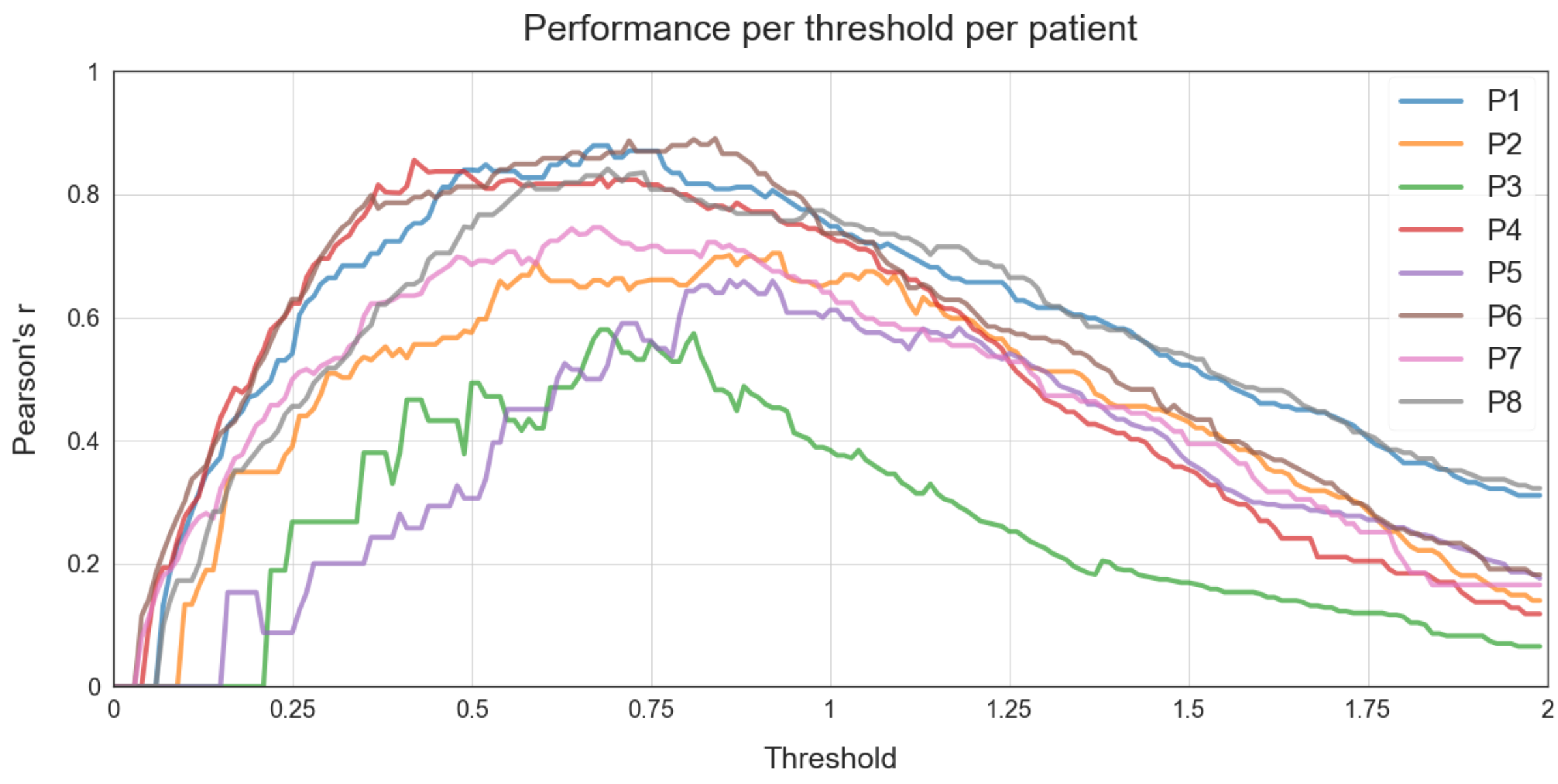

*Figure 6. Performance (y-axis) per threshold (x-axis) per patient individually using Pearson's r as a metric. Y-axis = closer to 1 the better the performance*

Table 3. Thresholds used for each version of NUVA (i) 'fixed' and (ii) 'adapted'.

| **System Version** | **fixed** | **adapted** | | | | | | | |
|---|---|---|---|---|---|---|---|---|---|
| **Data** | All Patients | P1 | P2 | P3 | P4 | P5 | P6 | P7 | P8 |
| **Best Threshold** | 0.680 | 0.670 | 0.920 | 0.680 | 0.420 | 0.860 | 0.840 | 0.670 | 0.690 |

## 4.2 Performance across all patients

Overall performance of NUVA for both versions, 'fixed' and 'adapted', are compared to a commercial baseline (Google STT) and a second independent SLT scorer, SLT2. Results are presented in Table 4Table 4.

NUVA yields better performance than the baseline for all metrics, except for False Positives rates. These differences are significant with respect to the baseline, as shown in Table 5. Notice that we did not find a significant difference in performance between the 'fixed' and 'adapted' version of NUVA.

In general, since scoring criteria between SLTs, which conforms the 'gold-standard', might differ slightly, we included the performance of a second independent SLT scorer (SLT2) with respect to the primary SLT scorer used in this study to illustrate the difference between the performance of all three computerised systems, and human SLT scores. Both baseline and NUVA, are significantly inferior to the performance of the second independent SLT scorer.

We ran Cochran's Q test (Cochran, 1950) to determine if the binary predictions of baseline, 'fixed' and 'adapted' versions of NUVA and also the SLT2 binary scores were statistically significantly different across all patients ($p<0.001$). We then performed pairwise comparisons using Dunn's procedure (Dunn, 1964) with Holm correction (Holm, 1979) for multiple comparisons, see Table 5.

Table 4. Performance across all patients of (i) baseline (Google STT service), (ii) 'fixed' version of NUVA (iii) 'adapted' version of NUVA, (iv) a second independent SLT, SLT2, against the primary SLT scorer used in this study.

| System | Accuracy | False Positives | False Negatives | F1-Score | Pearson's r |
|---|---|---|---|---|---|
| **baseline** | 0.882 | 0.020 | 0.098 | 0.795 | 0.727 |
| **fixed** | 0.905 | 0.049 | 0.046 | 0.855 | 0.784 |
| **adapted** | 0.913 | 0.057 | 0.031 | 0.871 | 0.807 |
| **SLT2** | 0.965 | 0.016 | 0.018 | 0.947 | 0.921 |

Table 5. Significance testing across all patients (as a group) using Dunn's post hoc test (with Holm correction) between systems' predictions and scores of a second independent SLT, SLT2. Notation: * $p < 0.05$, ** $p < 0.01$, *** $p < 0.001$ and NS, non-significant.

| Pairs | All Patients |
|---|---|
| **baseline-fixed** | * |
| **baseline-adapted** | ** |
| **baseline-SLT2** | *** |
| **fixed-adapted** | NS |
| **fixed-SLT2** | *** |
| **adapted-SLT2** | *** |

## 4.3 Performance per Patient

Each system's performance per patient is illustrated in Figure 7, together with a second independent SLT scorer, SLT2 (see Appendix A for details). Significance testing follows the same methodology from the previous section but applied to each patient. See significance testing in Table 6.

At least one version of NUVA had better performance than the baseline system for 7/8 patients, but not patient P4. These differences were statistically significant for 3/8 patients, see P6, P7 and P8. There was no significant difference between the 'fixed' and 'adapted' versions of NUVA. Performance of NUVA and SLT2 were not significantly different for 5/8 patients. NUVA performed worse for patients P2, P3 and P5.

In terms of false positives and false negatives, NUVA tended to have less false negatives than false positives (Figure 8). The baseline system, in contrast, had less false positives and more false negatives.

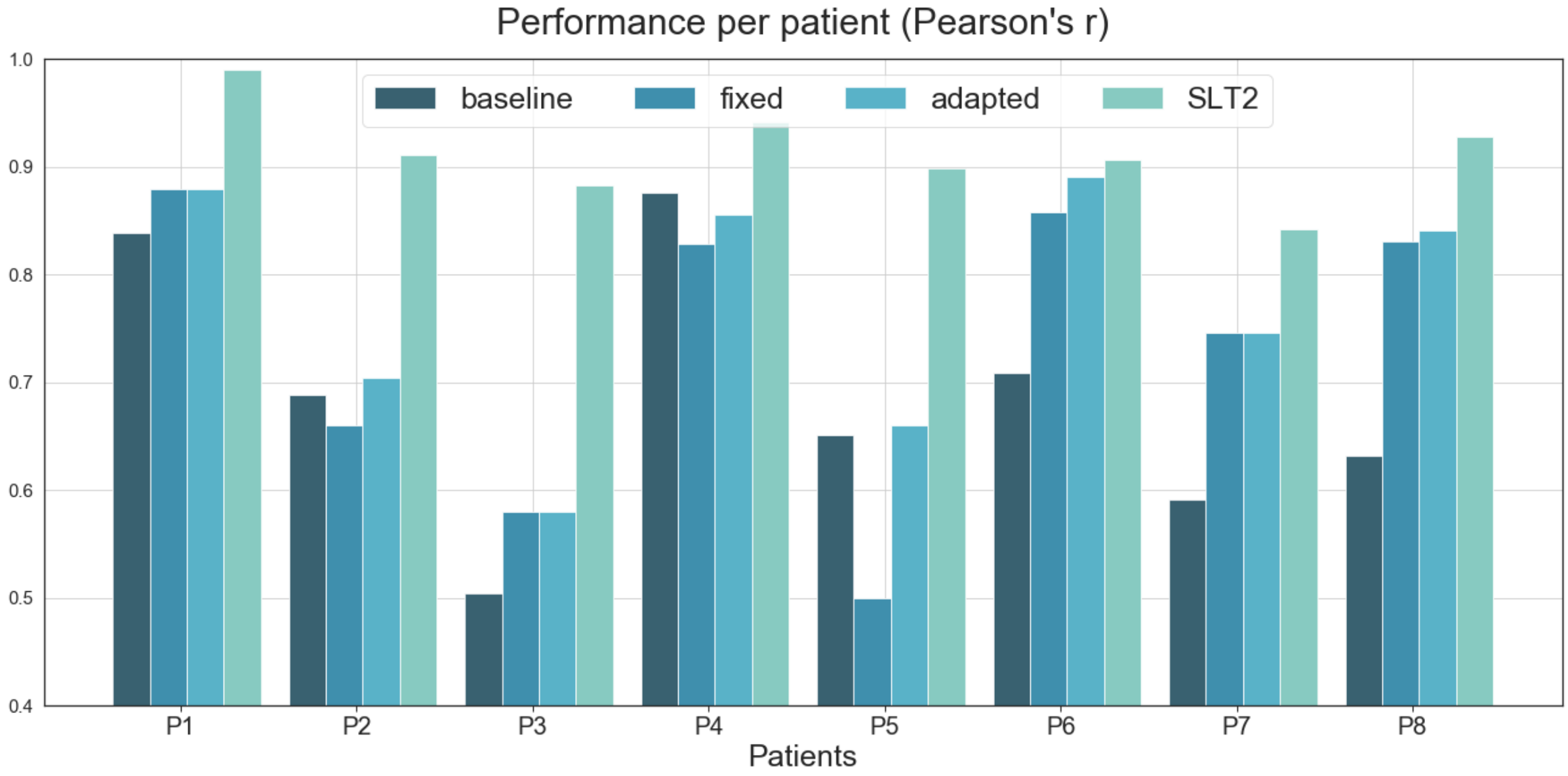

*Figure 7. Comparison of performance between (i) a commercial baseline (ii) the 'fixed' version of NUVA, (iii) the 'adapted' version of NUVA (iv) a second independent SLT, SLT2, against the primary SLT used in this study. Y-axis = closer to 1, the better the performance.*

Table 6. Significance testing per patient using Dunn's post hoc test (with Holm correction) between systems' predictions and scores of a second independent SLT, SLT2. Notation: * $p < 0.05$, ** $p < 0.01$, *** $p < 0.001$ and NS, non-significant.

| Pairs | P1 | P2 | P3 | P4 | P5 | P6 | P7 | P8 |
|---|---|---|---|---|---|---|---|---|
| **baseline-fixed** | NS | NS | NS | NS | NS | * | * | ** |
| **baseline-adapted** | NS | NS | NS | NS | NS | ** | * | ** |
| **baseline-SLT2** | ** | ** | NS | NS | NS | ** | *** | *** |
| **fixed-adapted** | NS | NS | NS | NS | NS | NS | NS | NS |
| **fixed-SLT2** | NS | ** | * | NS | ** | NS | NS | NS |
| **adapted-SLT2** | NS | ** | * | NS | * | NS | NS | NS |

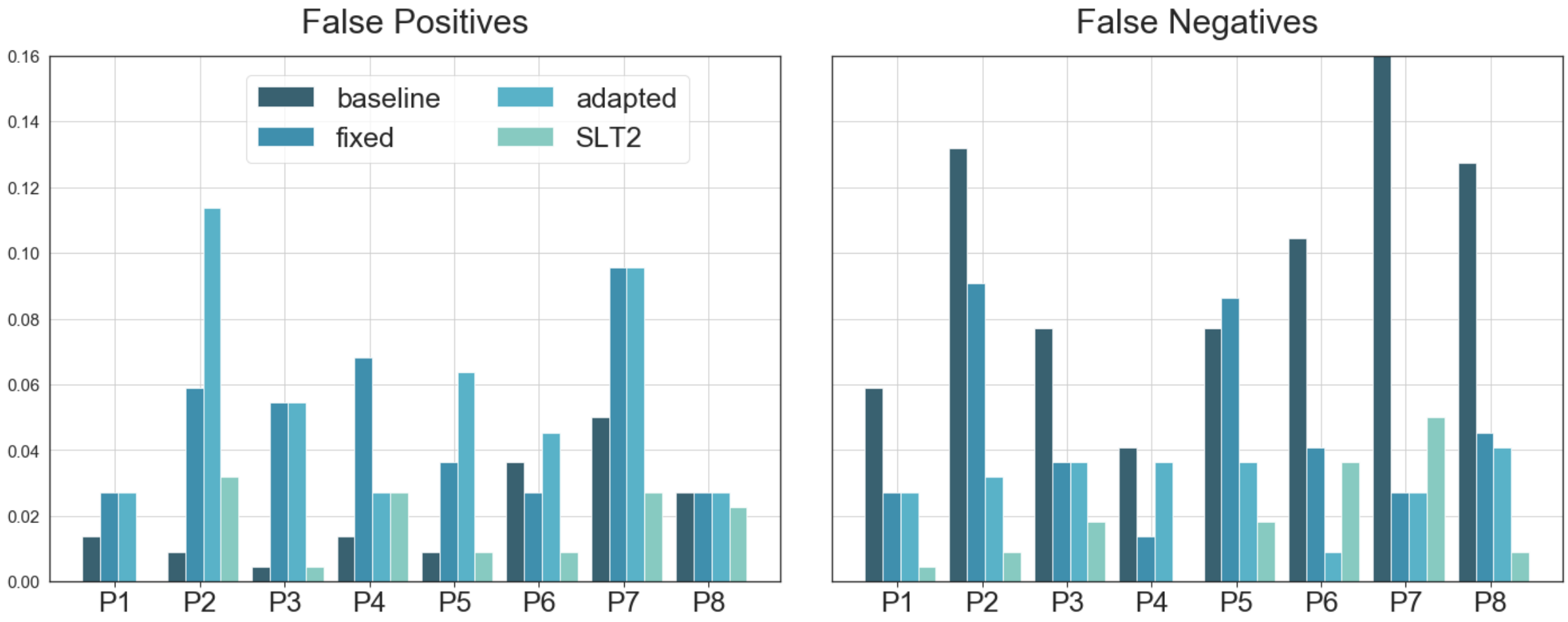

*Figure 8. False Positives and False Positives for (i) baseline, (ii) 'fixed' version of NUVA, (iii) 'adapted' version of NUVA and (iv) a second independent SLT, SLT2. Y-axis = less is better.*

## 4.4 Extent of agreement between systems and SLTs

To ease the visualisation of results we used a numerical coding system to report the extent of agreement with 95% certainty using Gwet's benchmarking method (Gwet, 2014) using the range proposed by McHugh for healthcare and clinical research (McHugh, 2012) , see Table 7Table 7 and Table 8. Calculations to obtain Table 8 can be found in Appendix B.2 using inter-rater reliability measures from Appendix B.1.

*Table 7. Benchmark range proposed by McHugh as agreement interpretation for applications in health care and clinical research. Code column is used as a short reference to the agreement description.*

| Benchmark Range | Description | Code |
|---|---|---|
| **above .90** | Almost Perfect | 6 |
| **.80-.90** | Strong | 5 |
| **.60-.80** | Moderate | 4 |
| **.40-.59** | Weak | 3 |
| **.21-.39** | Minimal | 2 |
| **below .20** | None | 1 |

*Table 8. Extend of agreement between systems and each SLT and, also between SLTs themselves at the bottom, calculated with a 95% certainty using Gwet's benchmarking method (Gwet, 2014) .*

| Agreement between | P1 | P2 | P3 | P4 | P5 | P6 | P7 | P8 | All |
|---|---|---|---|---|---|---|---|---|---|
| Baseline & SLT1 | 5 | 4 | 5 | 5 | 5 | 4 | 3 | 4 | 4 |
| Baseline & SLT2 | 5 | 4 | 5 | 5 | 5 | 4 | 3 | 4 | 4 |
| Fixed & SLT1 | 5 | 4 | 5 | 5 | 5 | **5** | **4** | 5 | 5 |
| Fixed & SLT2 | 5 | 4 | 5 | 5 | 4 | **5** | **4** | 5 | 4 |
| Adapted & SLT1 | 5 | 4 | 5 | 4 | 4 | **5** | **4** | 5 | 5 |
| Adapted & SLT2 | 5 | 4 | 5 | 5 | 5 | **5** | **4** | 5 | 4 |
| **SLT1 & SLT2** | **6** | **5** | **6** | **6** | **6** | **5** | **4** | **6** | **6** |

Across patients, none of the ASR systems reached human SLT level of agreement. However, 'fixed' and 'adapted' versions of NUVA reached the closest i.e., just one level below SLT1, the 'gold-standard' in this study, and two levels below SLT2. The baseline ASR system reached two levels below SLT expert regardless the SLT used.

In general, the 'fixed' version of NUVA reached closer agreement levels to SLTs per patient using the 'gold-standard' SLT1 and, the 'adapted' version using SLT2. However, this difference is only evident in patients 4 and 5.

## 4.5 System Cross-validation

To assess the predictive performance of NUVA and to judge how it performs outside the sample to a new unseen dataset, we used cross-validation. The assumption, in this case, was that previously collected speech samples from patients could be used to optimise the system's classifying threshold. This assumption is consistent with current practices as a patient always undertakes an initial assessment with an SLT before starting therapy; this assessment could provide the speech samples needed to calibrate NUVA's threshold. For each patient, a 10-fold cross-validation procedure was applied, and the average performance across folds is reported, together with minimum, maximum and range, see Table 9. Accuracies for all patients was high, above 84% with a range of 10% and a group average of 89.5%

Table 9. Results for a 10-fold cross-validation for each patient of the 'adapted' version of NUVA. For each patient, the average across all folds is reported as Mean (±SD).

| Patient | Accuracy | False Positives | False Negatives | F1-Score | Pearson's r |
|---|---|---|---|---|---|
| **P1** | 0.93(±0.068) | 0.04(±0.045) | 0.03(±0.046) | 0.89(±0.106) | 0.85(±0.149) |
| **P2** | 0.84(±0.082) | 0.11(±0.074) | 0.05(±0.052) | 0.78(±0.116) | 0.67(±0.162) |
| **P3** | 0.88(±0.055) | 0.08(±0.029) | 0.04(±0.038) | 0.51(±0.247) | 0.46(±0.278) |
| **P4** | 0.94(±0.055) | 0.03(±0.030) | 0.04(±0.040) | 0.89(±0.088) | 0.85(±0.123) |
| **P5** | 0.87(±0.060) | 0.09(±0.059) | 0.04(±0.032) | 0.61(±0.247) | 0.56(±0.261) |
| **P6** | 0.93(±0.071) | 0.05(±0.057) | 0.03(±0.030) | 0.91(±0.104) | 0.85(±0.150) |
| **P7** | 0.87(±0.081) | 0.10(±0.059) | 0.03(±0.041) | 0.90(±0.065) | 0.72(±0.183) |
| **P8** | 0.90(±0.038) | 0.05(±0.041) | 0.05(±0.043) | 0.85(±0.067) | 0.79(±0.087) |
| **Mean(SD)** | 0.895(0.03) | 0.066(0.03) | 0.039(0.01) | 0.790(0.14) | 0.718(0.14) |
| **Min** | 0.836 | 0.027 | 0.027 | 0.506 | 0.462 |
| **Max** | 0.936 | 0.114 | 0.05 | 0.905 | 0.852 |
| **Range** | 0.1 | 0.086 | 0.023 | 0.399 | 0.389 |

## 4.5 Latency in systems' response

We assessed the baseline and NUVA's performance in terms of speed to label 'correct'/'incorrect' a given naming attempt and speed in processing a second of speech. We used a Samsung tablet, SM-T820 model, running on Android 7 (SDK 24) to test NUVA's performance, as it was adapted to work embedded on a mobile platform. The baseline (Google Cloud Platform Speech-to-Text service) used a client-server model with gRPC protocol. Figure 9 and Table 10 illustrate for both systems (a) the speed to process 1760 naming recordings and (b) the average time to process one second of speech. In total, the baseline system had 37 outliers above 2.5 seconds. These have been clamped to 2.5 seconds in the visualisation but not in the calculations.

Table 10. Statistics of timing measurements (latency) of (i) baseline and (ii) NUVA in providing feedback. (a) Time to process each of the naming attempts in our test. (b) Average time to process a second of speech. The less, the better.

(a)

| Time (s) | baseline | NUVA |
|---|---|---|
| **Mean(±SD)** | 0.825(±0.362) | 0.605(±0.170) |
| **Min** | 0.481 | 0.227 |
| **Max** | 10.659 | 2.207 |
| **Range** | 10.178 | 1.980 |

(b)

| Time (s) | baseline | NUVA |
|---|---|---|
| **Mean(±SD)** | 0.571(±0.255) | 0.404(±0.058) |
| **Min** | 0.247 | 0.252 |
| **Max** | 6.513 | 0.851 |
| **Range** | 6.266 | 0.599 |

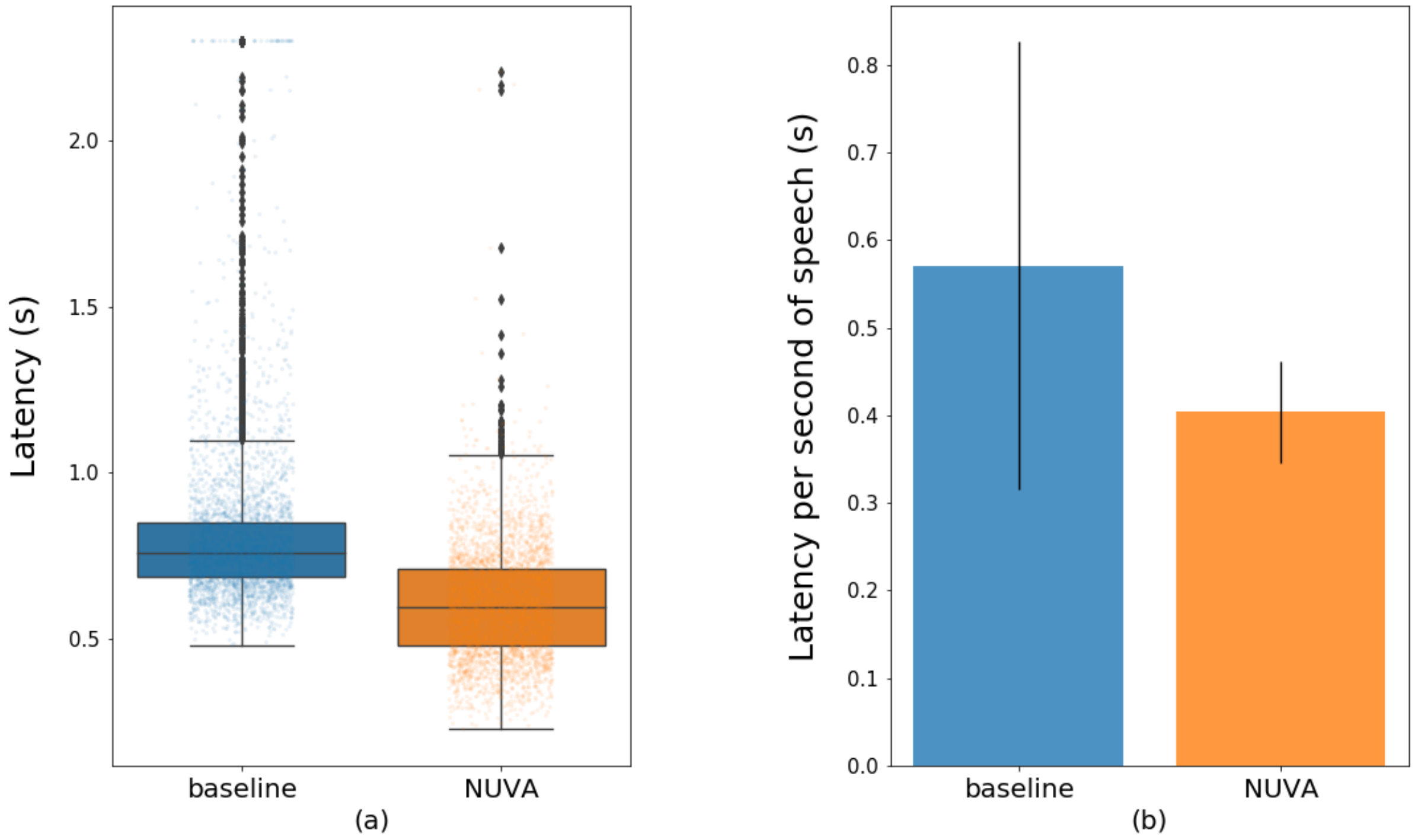

*Figure 9. Timing measurements (latency) in providing labels' correct'/'incorrect' for the baseline (Google STT service) and NUVA (embedded). (a) Time in seconds to process each of the naming attempts in our test. (b) Average time in seconds to process a second of speech. Y=Time in seconds, less is better.*

Using a paired Wilcoxon Signed Rank Test, performance speed was statistically significantly different between the two systems ($p<0.001$). On average NUVA was 1.42 times faster than the baseline system in processing each naming attempt. Since naming attempts in the test set varied in length, this translated in NUVA taking 0.404 seconds on average to process a second of speech while the baseline took 0.571 seconds.

## 5. Discussion

This study aimed to evaluate the effectiveness of NUVA, a tailor-made ASR system to assess spoken picture naming performance in a group of eight chronic aphasic stroke patients. Results show that NUVA had within sample performance accuracies above 85% across all patients, irrespective of the thresholding approach used, and reached comparable agreement with the therapist's ratings in two of the patients. A cross-validation analysis found accuracy remained high, above 84% for all patients, with an average of 89.5% overall. This demonstrates the feasibility of NUVA to reliably discriminate 'correct' from 'incorrect' spoken naming attempts in aphasic speech.

### 5.1 Baseline vs. NUVA

To offer the reader a benchmark as to what level of performance can be achieved with aphasic speech data using standard, 'off-the-shelf', publicly available STT ASR systems we compared the performance of our tailor-made ASR system NUVA to a commercial baseline (Google Cloud Platform STT service).

NUVA performed significantly better overall in terms of both accuracy and speed. Furthermore, only the NUVA system reached human expert SLT level of performance agreement, albeit only in two patients. Taken together these data demonstrate that a tailor-made ASR technology for aphasic speech such as NUVA is feasible and may be an effective solution for accurate, reliable and fast automatic, digitally delivered speech feedback system for aphasia interventions targeting spoken word naming performance.

#### *5.1.1 Accuracy*

In terms of accuracy, NUVA performed well overall when compared to the SLT 'gold standard'. The three patients (2, 3 and 5) where NUVA performed significantly worse with respect SLT2 scorings (Table 9), were the more severely speech impaired, *i.e.,* the lowest performers on the Object Naming subtest of the Comprehensive Aphasia Test (Swinburn et al., 2004). These results show that NUVA while promising, needs further improvement to reach a comparable human/SLT level in more severely impaired aphasic speakers. Similar performance decay is displayed by the baseline in patients 7 and 8 in addition to 2, 3 and 5, see Appendix A.

The same recordings were used by the baseline and NUVA systems and, both systems used British English corpora for training, however only the NUVA system was designed to accommodate the speech profile of PWA. This highlights both the need and potential to use tailor-made ASR technology to reach expert human level performance, (SLT), when processing speech data from clinical populations with aphasia.

#### *5.1.2 Agreement between systems and SLTs*

Percentage agreement and inter-rater reliability, as indexed by Gwet's AC1, was consistently high among SLTs reaching "Almost Perfect" levels of agreement for 5/8 patients. While patients 2, 6 and 7, yielded "Strong", "Strong" and "Moderate" agreement levels, respectively.

The NUVA system, 'fixed' or 'adapted', reached the same level of agreement as the SLTs for patients 6 and 7 with 95% certainty. In contrast the baseline system did not reach human SLT agreement levels for any patient. Taken together these results, albeit in a small patient sample suggest creating tailor-made ASR technology specifically for aphasic speech has the potential to reach human expert SLT level of performance on naming tasks.

#### *5.1.3 Latency on Feedback*

An essential feature of an automated ASR system for verifying spoken naming attempts is its ability to provide a timely response online, akin to an online SLT rater. When such as system is to be used in conjunction with a treatment intervention e.g., an anomia treatment program for PWA the speed in providing an automatic answer has direct implications in the efficiency of the treatment design. For example, the faster the ASR performance and feedback to the PWA the greater the number of items that can be treated within a given time window allowing in turn more fine-grained experimental investigation of the impact of dose and intensity on speech production recovery.

We assessed latency response times for both the baseline and NUVA systems to illustrate what level of performance is currently possible with 'off-the-shelf' ASR software and our tailor-made tool. In this regard there are significant differences between both systems: the baseline relies on a client-server model with lags due to network communications, NUVA is embedded and its performance depends on the hardware it is deployed into. Nevertheless, NUVA the utterance verifier in a mobile device was significantly faster in providing feedback for each naming attempt than the client-server model used by Google STT service powered by its high-performance RPC protocol, gRPC. On average, NUVA processed each naming attempt 1.42 times faster than the baseline system. NUVA had an average latency of 0.404 seconds for a second of speech resulting in latencies of 0.808, 1.01 and 1.212 seconds to process naming attempts of lengths 2, 2.5 and 3 seconds long respectively; the baseline, 1.14, 1.425 and 1.71 seconds respectively. Future work will investigate whether (i) these NUVA latency results are acceptable when deployed as a real-time performance feedback system in PWA and (ii) performance speed can be optimised further using quantization techniques for training and inference or simply acquiring more powerful hardware.

## 5.2 NUVA Performance Accuracy

In the literature, to date, two key studies have evaluated ASR systems' performance on aphasic speakers' word naming attempts (Abad et al., 2013; Ballard et al., 2019). The level of heterogeneity between the three studies (including ours) is high (different languages spoken, types of aphasia, level of impairment, vocabulary assessed) nevertheless, NUVA does appear to offer both a more accurate and less variable performance. Abad and Ballard report a mean accuracy performance of their systems of 82% (range: 24%) and 75% (range: 18%) respectively. In contrast, NUVA reached a performance of 89.5% with a range of just 10% across patients (see Table 11 for details). However, Ballard's results are arguably more reliable as their system's performance is assessed online.

*Table 11. Comparison of NUVA results with two key studies in the literature that have used ASR to assess spoken naming attempts in aphasia.*

| Study | Abad et al. 2013 | Ballard et al. 2019 | NUVA |
|---|---|---|---|
| **Language** | Portuguese | Australian English | British English |
| **Population** | Aphasia | Aphasia/Apraxia | Aphasia |
| **N** | 8 | 5 | 8 |
| **Stimuli** | 103 | 124 | 220 |
| **Word Productions** | 824 | 7069 | 1760 |
| **Interrater Reliability Agreement (method)** | >0.85 (point-to-point agreement) | 0.96 (point-to-point agreement) | 0.938 (Gwet's AC1) |
| **Accuracy** | 0.820 | 0.75 | **0.895** |
| **Min** | 0.690 | 0.651 | **0.836** |
| **Max** | 0.930 | 0.828 | **0.936** |
| **Range** | 0.240 | 0.177 | **0.100** |
| **Evaluation Method** | 10-Fold CV | Online | 10-Fold CV |

## 5.3 Limitations

Our results, including the high cross-validation performance data, while encouraging, are perhaps best seen as a proof of concept at this time. The patient sample is small, and further validation in a larger sample of patients is required. To address this, NUVA is being used within a clinical trial of a novel anomia treatment App - iTALKbetter. The aim here is for NUVA to assess online 'live' naming attempts so that aphasic patients' speech performance can be used to drive therapy progression, *e.g.,* when a patient names an item reliably three times, they no longer need to practice it, and they progress to practising new words and/or the ASR gives live feedback on a trial by trial basis to the patient and encourages their ongoing participation. In this context, any mistake of our system, *i.e.,* a misclassification of speech attempts, could affect a patient's recovery. False positives could reinforce error-full responses when actual 'incorrect' attempts are understood as 'correct', while false negatives could frustrate patients when their actual 'correct' attempts are misclassified as 'incorrect', possibly leading them to abandon the intervention. For NUVA, while 'adapted' version gave the best accuracy performance, the 'fixed' version had on average a lower rate of false positives. As such, this is the solution we have deployed in the clinical trial of iTALKbetter. Nevertheless, we are mindful that for individual patients, there was variance in the range of false negatives (e.g., P2 and P5 was 9%) and false positives (e.g., P7 -10%). Further qualitative testing will investigate if such misclassification rates are acceptable or not to people with aphasia and clinicians.

### 5.4 Next steps

A significant benefit in the approach we used to construct NUVA is that it relies on recordings of the target word by healthy subjects. Healthy speech is considerably easier to obtain than aphasic speech meaning vocabulary of target words (as long as we can find recordings for them) could be enlarged indefinitely without the need for retraining models. We also predict the system's performance can be improved by expanding the variety of healthy voices for each target word, *e.g.,* to include different accents.

Future steps will be oriented towards designing and developing a system that can offer a higher degree of information to a clinician other than just ‘correct’ or ‘incorrect’ performance. For example, if a PWA’s naming attempt was classified as incorrect, it may be helpful for the clinician to understand if there was a pattern to the PWA’s type of naming errors. Such as, did the PWA make a spoken naming attempt at all or were their errors mostly phonemically related to the target word e.g., /cap/ instead of /cat/ or semantically related e.g., /dog/.

### 6. Conclusions

We present here NUVA, a tailor-made utterance verification system based on a deep learning architecture to assess word naming attempts for people with aphasia automatically. In a sample of eight aphasic patients' 1760 naming attempts, NUVA performed significantly better than the commercial baseline (Google STT service). Given the scarcity of aphasic speech corpora, this represents a significant step towards creating a reliable and automatic spoken word assessment system for aphasic speakers. It also offers clinicians a deployable solution for gathering big data, optimisation of similar systems and further research.

Future work will focus on optimising the effectiveness of NUVA to deliver live (accurate and fast) speech performance feedback within a clinical trial using a digitally delivered anomia treatment App: iTALKbetter. NUVA is available open-source to encourage reproducibility and further development in this field; we welcome new insights and collaborations. Despite being initially trained in English, given the language-agnostic framework of our system, it would be interesting to see if it can be used in other languages. This would offer an invaluable tool for aphasic speakers of under-researched languages.

## 7. Acknowledgements

DB and the ASR technical development is funded by a Medical Research Council iCASE PhD studentship, award number 1803748, JC is supported by a Wellcome Trust Senior Research Fellowship in Clinical Science (106161/Z/14/Z), and APL, VF, EU, HCF and CD by an NIHR Research Professorship.

## Appendix A

Performance per patient of (i) baseline (Google STT), (ii) 'fixed' version of NUVA, (iii) 'adapted' version of NUVA and (iv) a second independent SLT scorer, SLT2, against the primary SLT scorer used in this study.

| P1 | Accuracy | False Positives | False Negatives | F1-Score | Pearson's r |
|---|---|---|---|---|---|
| **baseline** | 0.927 | 0.014 | 0.059 | 0.887 | 0.839 |
| **fixed** | 0.945 | 0.027 | 0.027 | 0.921 | 0.879 |
| **adapted** | 0.945 | 0.027 | 0.027 | 0.921 | 0.879 |
| **SLT2** | 0.995 | 0.000 | 0.005 | 0.993 | 0.990 |

| P5 | Accuracy | False Positives | False Negatives | F1-Score | Pearson's r |
|---|---|---|---|---|---|
| **baseline** | 0.914 | 0.009 | 0.077 | 0.667 | 0.651 |
| **fixed** | 0.877 | 0.036 | 0.086 | 0.557 | 0.500 |
| **adapted** | 0.900 | 0.064 | 0.036 | 0.718 | 0.660 |
| **SLT2** | 0.973 | 0.009 | 0.018 | 0.914 | 0.899 |

| P2 | Accuracy | False Positives | False Negatives | F1-Score | Pearson's r |
|---|---|---|---|---|---|
| **baseline** | 0.859 | 0.009 | 0.132 | 0.748 | 0.688 |
| **fixed** | 0.850 | 0.059 | 0.091 | 0.769 | 0.660 |
| **adapted** | 0.855 | 0.114 | 0.032 | 0.810 | 0.705 |
| **SLT2** | 0.959 | 0.032 | 0.009 | 0.942 | 0.912 |

| P6 | Accuracy | False Positives | False Negatives | F1-Score | Pearson's r |
|---|---|---|---|---|---|
| **baseline** | 0.859 | 0.036 | 0.105 | 0.812 | 0.708 |
| **fixed** | 0.932 | 0.027 | 0.041 | 0.915 | 0.859 |
| **adapted** | 0.945 | 0.045 | 0.009 | 0.936 | 0.891 |
| **SLT2** | 0.955 | 0.009 | 0.036 | 0.943 | 0.906 |

| P3 | Accuracy | False Positives | False Negatives | F1-Score | Pearson's r |
|---|---|---|---|---|---|
| **baseline** | 0.918 | 0.005 | 0.077 | 0.471 | 0.504 |
| **fixed** | 0.909 | 0.055 | 0.036 | 0.630 | 0.580 |
| **adapted** | 0.909 | 0.055 | 0.036 | 0.630 | 0.580 |
| **SLT2** | 0.977 | 0.005 | 0.018 | 0.894 | 0.883 |

| P7 | Accuracy | False Positives | False Negatives | F1-Score | Pearson's r |
|---|---|---|---|---|---|
| **baseline** | 0.786 | 0.050 | 0.164 | 0.802 | 0.591 |
| **fixed** | 0.877 | 0.095 | 0.027 | 0.903 | 0.746 |
| **adapted** | 0.877 | 0.095 | 0.027 | 0.903 | 0.746 |
| **SLT2** | 0.923 | 0.027 | 0.050 | 0.934 | 0.842 |

| P4 | Accuracy | False Positives | False Negatives | F1-Score | Pearson's r |
|---|---|---|---|---|---|
| **baseline** | 0.945 | 0.014 | 0.041 | 0.914 | 0.876 |
| **fixed** | 0.918 | 0.068 | 0.014 | 0.886 | 0.829 |
| **adapted** | 0.936 | 0.027 | 0.036 | 0.903 | 0.856 |
| **SLT2** | 0.973 | 0.027 | 0.000 | 0.961 | 0.941 |

| P8 | Accuracy | False Positives | False Negatives | F1-Score | Pearson's r |
|---|---|---|---|---|---|
| **baseline** | 0.845 | 0.027 | 0.127 | 0.712 | 0.632 |
| **fixed** | 0.927 | 0.027 | 0.045 | 0.882 | 0.831 |
| **adapted** | 0.932 | 0.027 | 0.041 | 0.891 | 0.841 |
| **SLT2** | 0.968 | 0.023 | 0.009 | 0.951 | 0.928 |

# Appendix B.1

Measures of inter-rater reliability, percentage agreement (**PA**) and Gwet's AC1 agreement coefficient, together with standard error (**StdErr**) and 95% confidence interval (**CI**) between systems and each speech and language therapist (SLT1 and SLT2).

**Baseline & SLT1**

| Patient | PA | Gwet's AC1 | StdErr | 95% CI |
|---|---|---|---|---|
| P1 | 92.73% | 0.871 | 0.031 | (0.810, 0.932) |
| P2 | 85.91% | 0.764 | 0.039 | (0.687, 0.841) |
| P3 | 91.82% | 0.905 | 0.022 | (0.862, 0.947) |
| P4 | 94.55% | 0.904 | 0.027 | (0.851, 0.957) |
| P5 | 91.36% | 0.888 | 0.024 | (0.841, 0.936) |
| P6 | 85.91% | 0.735 | 0.044 | (0.648, 0.821) |
| P7 | 78.64% | 0.575 | 0.055 | (0.468, 0.683) |
| P8 | 84.55% | 0.746 | 0.040 | (0.667, 0.824) |
| All | 88.18% | 0.799 | 0.013 | (0.774, 0.825) |

**Fixed & SLT2**

| Patient | PA | Gwet's AC1 | StdErr | 95% CI |
|---|---|---|---|---|
| P1 | 94.09% | 0.892 | 0.029 | (0.836, 0.949) |
| P2 | 84.09% | 0.696 | 0.047 | (0.603, 0.788) |
| P3 | 89.55% | 0.869 | 0.026 | (0.818, 0.919) |
| P4 | 92.73% | 0.868 | 0.032 | (0.806, 0.930) |
| P5 | 89.09% | 0.847 | 0.029 | (0.790, 0.905) |
| P6 | 92.73% | 0.859 | 0.034 | (0.792, 0.925) |
| P7 | 83.64% | 0.690 | 0.047 | (0.597, 0.783) |
| P8 | 92.73% | 0.872 | 0.031 | (0.811, 0.932) |
| All | 89.83% | 0.816 | 0.013 | (0.790, 0.841) |

**Baseline & SLT2**

| Patient | PA | Gwet's AC1 | StdErr | 95% CI |
|---|---|---|---|---|
| P1 | 92.27% | 0.863 | 0.032 | (0.801, 0.926) |
| P2 | 84.55% | 0.737 | 0.041 | (0.656, 0.818) |
| P3 | 93.18% | 0.922 | 0.020 | (0.883, 0.960) |
| P4 | 93.64% | 0.886 | 0.030 | (0.828, 0.944) |
| P5 | 91.36% | 0.889 | 0.024 | (0.842, 0.937) |
| P6 | 88.64% | 0.789 | 0.040 | (0.711, 0.867) |
| P7 | 79.09% | 0.583 | 0.055 | (0.476, 0.690) |
| P8 | 86.82% | 0.781 | 0.038 | (0.706, 0.855) |
| All | 88.69% | 0.808 | 0.013 | (0.783, 0.833) |

**Adapted & SLT1**

| Patient | PA | Gwet's AC1 | StdErr | 95% CI |
|---|---|---|---|---|
| P1 | 94.55% | 0.900 | 0.028 | (0.846, 0.955) |
| P2 | 85.00% | 0.733 | 0.043 | (0.649, 0.817) |
| P3 | 90.91% | 0.884 | 0.025 | (0.836, 0.933) |
| P4 | 91.82% | 0.848 | 0.034 | (0.781, 0.916) |
| P5 | 87.73% | 0.839 | 0.029 | (0.782, 0.896) |
| P6 | 93.18% | 0.869 | 0.033 | (0.804, 0.933) |
| P7 | 87.73% | 0.770 | 0.041 | (0.689, 0.851) |
| P8 | 92.73% | 0.873 | 0.031 | (0.813, 0.933) |
| All | 90.45% | 0.829 | 0.013 | (0.805, 0.854) |

**Fixed & SLT1**

| Patient | PA | Gwet's AC1 | StdErr | 95% CI |
|---|---|---|---|---|
| P1 | 94.55% | 0.900 | 0.028 | (0.846, 0.955) |
| P2 | 85.45% | 0.724 | 0.045 | (0.636, 0.813) |
| P3 | 90.91% | 0.884 | 0.025 | (0.836, 0.933) |
| P4 | 93.64% | 0.886 | 0.029 | (0.829, 0.944) |
| P5 | 90.00% | 0.859 | 0.029 | (0.803, 0.915) |
| P6 | 94.55% | 0.893 | 0.030 | (0.834, 0.952) |
| P7 | 87.73% | 0.770 | 0.041 | (0.689, 0.851) |
| P8 | 93.18% | 0.881 | 0.030 | (0.822, 0.939) |
| All | 91.25% | 0.841 | 0.012 | (0.817, 0.865) |

**Adapted & SLT2**

| Patient | PA | Gwet's AC1 | StdErr | 95% CI |
|---|---|---|---|---|
| P1 | 94.09% | 0.892 | 0.029 | (0.836, 0.949) |
| P2 | 84.55% | 0.721 | 0.044 | (0.635, 0.807) |
| P3 | 89.55% | 0.869 | 0.026 | (0.818, 0.919) |
| P4 | 92.73% | 0.863 | 0.033 | (0.799, 0.928) |
| P5 | 88.64% | 0.852 | 0.028 | (0.797, 0.907) |
| P6 | 93.18% | 0.870 | 0.032 | (0.807, 0.934) |
| P7 | 83.64% | 0.690 | 0.047 | (0.597, 0.783) |
| P8 | 92.27% | 0.864 | 0.032 | (0.802, 0.926) |
| All | 89.83% | 0.818 | 0.013 | (0.793, 0.843) |

## Appendix B.2

Agreement between systems and each speech and language therapist (SLT1 and SLT2) using Gwet's benchmarking method (Gwet, 2014) on McHugh's benchmark range (McHugh, 2012).

Interval Membership Probability is defined as $IMP = P\left(\frac{K_G - b}{SE} \leq Z \leq \frac{K_G - a}{SE}\right)$, where $K_G$ is Gwet's AC1, $SE$ is the standard error and interval $(a, b)$ determine an agreement level in a benchmark range. See Gwet's AC1 and standard errors between systems and each SLT in Appendix B.1.

| | **Benchmark** | | | **Cumulative membership probabilities** | | | | | | | | | **Membership probabilities** | | | | | | | | |
|---|---|---|---|---|---|---|---|---|---|---|---|---|---|---|---|---|---|---|---|---|---|
| | **Range** | **Description** | **Code** | **P1** | **P2** | **P3** | **P4** | **P5** | **P6** | **P7** | **P8** | **All** | **P1** | **P2** | **P3** | **P4** | **P5** | **P6** | **P7** | **P8** | **All** |
| **SLT1 & SLT2** | above .90 | Almost Perfect | 6 | **1.000** | 0.843 | **1.000** | **0.994** | **1.000** | 0.684 | 0.075 | **0.980** | **1.000** | 1.000 | 0.843 | 1.000 | 0.994 | 1.000 | 0.684 | 0.075 | 0.980 | 1.000 |
| | .80-.90 | Strong | 5 | 1.000 | **1.000** | 1.000 | 1.000 | 1.000 | **1.000** | 0.922 | 1.000 | 1.000 | 0.000 | 0.157 | 0.000 | 0.006 | 0.000 | 0.316 | 0.847 | 0.020 | 0.000 |
| | .60-.80 | Moderate | 4 | 1.000 | 1.000 | 1.000 | 1.000 | 1.000 | 1.000 | **1.000** | 1.000 | 1.000 | 0.000 | 0.000 | 0.000 | 0.000 | 0.000 | 0.000 | 0.078 | 0.000 | 0.000 |
| | .40-.59 | Weak | 3 | 1.000 | 1.000 | 1.000 | 1.000 | 1.000 | 1.000 | 1.000 | 1.000 | 1.000 | 0.000 | 0.000 | 0.000 | 0.000 | 0.000 | 0.000 | 0.000 | 0.000 | 0.000 |
| | .21-.39 | Minimal | 2 | 1.000 | 1.000 | 1.000 | 1.000 | 1.000 | 1.000 | 1.000 | 1.000 | 1.000 | 0.000 | 0.000 | 0.000 | 0.000 | 0.000 | 0.000 | 0.000 | 0.000 | 0.000 |
| | below .20 | None | 1 | 1.000 | 1.000 | 1.000 | 1.000 | 1.000 | 1.000 | 1.000 | 1.000 | 1.000 | 0.000 | 0.000 | 0.000 | 0.000 | 0.000 | 0.000 | 0.000 | 0.000 | 0.000 |
| | **Agreement with 95% certainty:** | | | **6** | **5** | **6** | **6** | **6** | **5** | **4** | **6** | **6** | | | | | | | | | |

| | **Benchmark** | | | **Cumulative Membership Probabilities** | | | | | | | | | **Interval Membership Probabilities** | | | | | | | | |
|---|---|---|---|---|---|---|---|---|---|---|---|---|---|---|---|---|---|---|---|---|---|
| | **Range** | **Description** | **Code** | **P1** | **P2** | **P3** | **P4** | **P5** | **P6** | **P7** | **P8** | **All** | **P1** | **P2** | **P3** | **P4** | **P5** | **P6** | **P7** | **P8** | **All** |
| **Baseline & SLT1** | above .90 | Almost Perfect | 6 | 0.174 | 0.000 | 0.584 | 0.554 | 0.319 | 0.000 | 0.000 | 0.000 | 0.000 | 0.174 | 0.000 | 0.584 | 0.554 | 0.319 | 0.000 | 0.000 | 0.000 | 0.000 |
| | .80-.90 | Strong | 5 | **0.989** | 0.180 | **1.000** | **1.000** | **1.000** | 0.070 | 0.000 | 0.088 | 0.484 | 0.815 | 0.180 | 0.416 | 0.446 | 0.681 | 0.070 | 0.000 | 0.087 | 0.484 |
| | .60-.80 | Moderate | 4 | 1.000 | **1.000** | 1.000 | 1.000 | 1.000 | **0.999** | 0.326 | **1.000** | **1.000** | 0.011 | 0.820 | 0.000 | 0.000 | 0.000 | 0.929 | 0.326 | 0.912 | 0.516 |
| | .40-.59 | Weak | 3 | 1.000 | 1.000 | 1.000 | 1.000 | 1.000 | 1.000 | **0.999** | 1.000 | 1.000 | 0.000 | 0.000 | 0.000 | 0.000 | 0.000 | 0.001 | 0.673 | 0.000 | 0.000 |
| | .21-.39 | Minimal | 2 | 1.000 | 1.000 | 1.000 | 1.000 | 1.000 | 1.000 | 1.000 | 1.000 | 1.000 | 0.000 | 0.000 | 0.000 | 0.000 | 0.000 | 0.000 | 0.001 | 0.000 | 0.000 |
| | below .20 | None | 1 | 1.000 | 1.000 | 1.000 | 1.000 | 1.000 | 1.000 | 1.000 | 1.000 | 1.000 | 0.000 | 0.000 | 0.000 | 0.000 | 0.000 | 0.000 | 0.000 | 0.000 | 0.000 |
| | **Agreement with 95% certainty:** | | | **5** | **4** | **5** | **5** | **5** | **4** | **3** | **4** | **4** | | | | | | | | | |

| | **Benchmark** | | | **Cumulative Membership Probabilities** | | | | | | | | | **Interval Membership Probabilities** | | | | | | | | |
|---|---|---|---|---|---|---|---|---|---|---|---|---|---|---|---|---|---|---|---|---|---|
| | **Range** | **Description** | **Code** | **P1** | **P2** | **P3** | **P4** | **P5** | **P6** | **P7** | **P8** | **All** | **P1** | **P2** | **P3** | **P4** | **P5** | **P6** | **P7** | **P8** | **All** |
| **Baseline & SLT2** | above .90 | Almost Perfect | 6 | 0.124 | 0.000 | 0.865 | 0.314 | 0.332 | 0.003 | 0.000 | 0.001 | 0.000 | 0.124 | 0.000 | 0.865 | 0.314 | 0.332 | 0.003 | 0.000 | 0.001 | 0.000 |
| | .80-.90 | Strong | 5 | **0.976** | 0.064 | **1.000** | **0.998** | **1.000** | 0.391 | 0.000 | 0.306 | 0.744 | 0.852 | 0.064 | 0.135 | 0.684 | 0.668 | 0.388 | 0.000 | 0.305 | 0.744 |
| | .60-.80 | Moderate | 4 | 1.000 | **1.000** | 1.000 | 1.000 | 1.000 | **1.000** | 0.378 | **1.000** | **1.000** | 0.024 | 0.935 | 0.000 | 0.002 | 0.000 | 0.609 | 0.378 | 0.694 | 0.256 |
| | .40-.59 | Weak | 3 | 1.000 | 1.000 | 1.000 | 1.000 | 1.000 | 1.000 | **1.000** | 1.000 | 1.000 | 0.000 | 0.000 | 0.000 | 0.000 | 0.000 | 0.000 | 0.621 | 0.000 | 0.000 |
| | .21-.39 | Minimal | 2 | 1.000 | 1.000 | 1.000 | 1.000 | 1.000 | 1.000 | 1.000 | 1.000 | 1.000 | 0.000 | 0.000 | 0.000 | 0.000 | 0.000 | 0.000 | 0.000 | 0.000 | 0.000 |
| | below .20 | None | 1 | 1.000 | 1.000 | 1.000 | 1.000 | 1.000 | 1.000 | 1.000 | 1.000 | 1.000 | 0.000 | 0.000 | 0.000 | 0.000 | 0.000 | 0.000 | 0.000 | 0.000 | 0.000 |
| | **Agreement with 95% certainty:** | | | **5** | **4** | **5** | **5** | **5** | **4** | **3** | **4** | **4** | | | | | | | | | |

| | Benchmark | | | Cumulative Membership Probabilities | | | | | | | | | Interval Membership Probabilities | | | | | | | | |
|---|---|---|---|---|---|---|---|---|---|---|---|---|---|---|---|---|---|---|---|---|---|
| | Range | Description | Code | P1 | P2 | P3 | P4 | P5 | P6 | P7 | P8 | All | P1 | P2 | P3 | P4 | P5 | P6 | P7 | P8 | All |
| **Fixed & SLT1** | above .90 | Almost Perfect | 6 | 0.506 | 0.000 | 0.260 | 0.321 | 0.075 | 0.410 | 0.001 | 0.257 | 0.000 | 0.506 | 0.000 | 0.260 | 0.321 | 0.075 | 0.410 | 0.001 | 0.257 | 0.000 |
| | .80-.90 | Strong | 5 | **1.000** | 0.047 | **1.000** | **0.998** | **0.980** | **0.999** | 0.235 | **0.997** | **1.000** | 0.494 | 0.047 | 0.739 | 0.678 | 0.906 | 0.589 | 0.234 | 0.739 | 1.000 |
| | .60-.80 | Moderate | 4 | 1.000 | **0.997** | 1.000 | 1.000 | 1.000 | 1.000 | **1.000** | 1.000 | 1.000 | 0.000 | 0.950 | 0.000 | 0.002 | 0.020 | 0.001 | 0.765 | 0.003 | 0.000 |
| | .40-.59 | Weak | 3 | 1.000 | 1.000 | 1.000 | 1.000 | 1.000 | 1.000 | 1.000 | 1.000 | 1.000 | 0.000 | 0.003 | 0.000 | 0.000 | 0.000 | 0.000 | 0.000 | 0.000 | 0.000 |
| | .21-.39 | Minimal | 2 | 1.000 | 1.000 | 1.000 | 1.000 | 1.000 | 1.000 | 1.000 | 1.000 | 1.000 | 0.000 | 0.000 | 0.000 | 0.000 | 0.000 | 0.000 | 0.000 | 0.000 | 0.000 |
| | below .20 | None | 1 | 1.000 | 1.000 | 1.000 | 1.000 | 1.000 | 1.000 | 1.000 | 1.000 | 1.000 | 0.000 | 0.000 | 0.000 | 0.000 | 0.000 | 0.000 | 0.000 | 0.000 | 0.000 |
| | **Agreement with 95% certainty:** | | | **5** | **4** | **5** | **5** | **5** | **5** | **4** | **5** | **5** | | | | | | | | | |

| | Benchmark | | | Cumulative Membership Probabilities | | | | | | | | | Interval Membership Probabilities | | | | | | | | |
|---|---|---|---|---|---|---|---|---|---|---|---|---|---|---|---|---|---|---|---|---|---|
| | Range | Description | Code | P1 | P2 | P3 | P4 | P5 | P6 | P7 | P8 | All | P1 | P2 | P3 | P4 | P5 | P6 | P7 | P8 | All |
| **Fixed & SLT2** | above .90 | Almost Perfect | 6 | 0.396 | 0.000 | 0.112 | 0.156 | 0.037 | 0.113 | 0.000 | 0.179 | 0.000 | 0.396 | 0.000 | 0.112 | 0.156 | 0.037 | 0.113 | 0.000 | 0.179 | 0.000 |
| | .80-.90 | Strong | 5 | **0.999** | 0.014 | **0.996** | **0.984** | 0.946 | **0.958** | 0.010 | **0.990** | 0.884 | 0.603 | 0.014 | 0.883 | 0.827 | 0.909 | 0.845 | 0.010 | 0.811 | 0.884 |
| | .60-.80 | Moderate | 4 | 1.000 | **0.979** | 1.000 | 1.000 | **1.000** | 1.000 | **0.972** | 1.000 | **1.000** | 0.001 | 0.965 | 0.004 | 0.016 | 0.054 | 0.042 | 0.962 | 0.010 | 0.116 |
| | .40-.59 | Weak | 3 | 1.000 | 1.000 | 1.000 | 1.000 | 1.000 | 1.000 | 1.000 | 1.000 | 1.000 | 0.000 | 0.021 | 0.000 | 0.000 | 0.000 | 0.000 | 0.028 | 0.000 | 0.000 |
| | .21-.39 | Minimal | 2 | 1.000 | 1.000 | 1.000 | 1.000 | 1.000 | 1.000 | 1.000 | 1.000 | 1.000 | 0.000 | 0.000 | 0.000 | 0.000 | 0.000 | 0.000 | 0.000 | 0.000 | 0.000 |
| | below .20 | None | 1 | 1.000 | 1.000 | 1.000 | 1.000 | 1.000 | 1.000 | 1.000 | 1.000 | 1.000 | 0.000 | 0.000 | 0.000 | 0.000 | 0.000 | 0.000 | 0.000 | 0.000 | 0.000 |
| | **Agreement with 95% certainty:** | | | **5** | **4** | **5** | **5** | **4** | **5** | **4** | **5** | **4** | | | | | | | | | |

| | Benchmark | | | Cumulative Membership Probabilities | | | | | | | | | Interval Membership Probabilities | | | | | | | | |
|---|---|---|---|---|---|---|---|---|---|---|---|---|---|---|---|---|---|---|---|---|---|
| | Range | Description | Code | P1 | P2 | P3 | P4 | P5 | P6 | P7 | P8 | All | P1 | P2 | P3 | P4 | P5 | P6 | P7 | P8 | All |
| **Adapted & SLT1** | above .90 | Almost Perfect | 6 | 0.506 | 0.000 | 0.260 | 0.066 | 0.018 | 0.169 | 0.001 | 0.189 | 0.000 | 0.506 | 0.000 | 0.260 | 0.066 | 0.018 | 0.169 | 0.001 | 0.189 | 0.000 |
| | .80-.90 | Strong | 5 | **1.000** | 0.058 | **1.000** | 0.921 | 0.909 | **0.982** | 0.235 | **0.992** | **0.990** | 0.494 | 0.058 | 0.739 | 0.855 | 0.891 | 0.813 | 0.234 | 0.803 | 0.990 |
| | .60-.80 | Moderate | 4 | 1.000 | **0.999** | 1.000 | **1.000** | **1.000** | 1.000 | **1.000** | 1.000 | 1.000 | 0.000 | 0.941 | 0.000 | 0.079 | 0.091 | 0.018 | 0.765 | 0.008 | 0.010 |
| | .40-.59 | Weak | 3 | 1.000 | 1.000 | 1.000 | 1.000 | 1.000 | 1.000 | 1.000 | 1.000 | 1.000 | 0.000 | 0.001 | 0.000 | 0.000 | 0.000 | 0.000 | 0.000 | 0.000 | 0.000 |
| | .21-.39 | Minimal | 2 | 1.000 | 1.000 | 1.000 | 1.000 | 1.000 | 1.000 | 1.000 | 1.000 | 1.000 | 0.000 | 0.000 | 0.000 | 0.000 | 0.000 | 0.000 | 0.000 | 0.000 | 0.000 |
| | below .20 | None | 1 | 1.000 | 1.000 | 1.000 | 1.000 | 1.000 | 1.000 | 1.000 | 1.000 | 1.000 | 0.000 | 0.000 | 0.000 | 0.000 | 0.000 | 0.000 | 0.000 | 0.000 | 0.000 |
| | **Agreement with 95% certainty:** | | | **5** | **4** | **5** | **4** | **4** | **5** | **4** | **5** | **5** | | | | | | | | | |

| | Benchmark | | | Cumulative Membership Probabilities | | | | | | | | | Interval Membership Probabilities | | | | | | | | |
|---|---|---|---|---|---|---|---|---|---|---|---|---|---|---|---|---|---|---|---|---|---|
| | Range | Description | Code | P1 | P2 | P3 | P4 | P5 | P6 | P7 | P8 | All | P1 | P2 | P3 | P4 | P5 | P6 | P7 | P8 | All |
| **Adapted & SLT2** | above .90 | Almost Perfect | 6 | 0.396 | 0.000 | 0.112 | 0.133 | 0.042 | 0.178 | 0.000 | 0.127 | 0.000 | 0.396 | 0.000 | 0.112 | 0.133 | 0.042 | 0.178 | 0.000 | 0.127 | 0.000 |
| | .80-.90 | Strong | 5 | **0.999** | 0.036 | **0.996** | **0.973** | **0.969** | **0.985** | 0.010 | **0.978** | 0.920 | 0.603 | 0.036 | 0.883 | 0.840 | 0.926 | 0.807 | 0.010 | 0.851 | 0.920 |
| | .60-.80 | Moderate | 4 | 1.000 | **0.997** | 1.000 | 1.000 | 1.000 | 1.000 | **0.972** | 1.000 | **1.000** | 0.001 | 0.961 | 0.004 | 0.027 | 0.031 | 0.015 | 0.962 | 0.022 | 0.080 |
| | .40-.59 | Weak | 3 | 1.000 | 1.000 | 1.000 | 1.000 | 1.000 | 1.000 | 1.000 | 1.000 | 1.000 | 0.000 | 0.003 | 0.000 | 0.000 | 0.000 | 0.000 | 0.028 | 0.000 | 0.000 |
| | .21-.39 | Minimal | 2 | 1.000 | 1.000 | 1.000 | 1.000 | 1.000 | 1.000 | 1.000 | 1.000 | 1.000 | 0.000 | 0.000 | 0.000 | 0.000 | 0.000 | 0.000 | 0.000 | 0.000 | 0.000 |
| | below .20 | None | 1 | 1.000 | 1.000 | 1.000 | 1.000 | 1.000 | 1.000 | 1.000 | 1.000 | 1.000 | 0.000 | 0.000 | 0.000 | 0.000 | 0.000 | 0.000 | 0.000 | 0.000 | 0.000 |
| | **Agreement with 95% certainty:** | | | **5** | **4** | **5** | **5** | **5** | **5** | **4** | **5** | **4** | | | | | | | | | |

# Appendix C

List of the 220 words selected as picture naming stimuli for this study based on the top 2000 most frequent words using the *Zipf* index of the SUBTLEX-UK corpus (van Heuven et al., 2014) keeping the same distribution of parts of speech for nouns, verbs, adverbs and adjectives. Included are the main descriptive attributes for each word: British English phonetic transcription based on the International Phonetic Alphabet (**British Pronunciation**) (*Cambridge Dictionary | English Dictionary, Translations & Thesaurus*, 2020), number of characters (**Ch**), number of phonemes in UK English (**Ph**), number of syllables (**Sy**), the *Zipf* index of the SUBTLEX-UK corpus (**Zipf**) (van Heuven et al., 2014), Part of Speech (**PoS**), Age of Acquisition (**AoA**) (Kuperman et al., 2012) and Concreteness (**Con.**) (Brysbaert et al., 2014).

| Word | British Pronunciation | Ch | Ph | Sy | Zipf | PoS | AoA | Con. |
|---|---|---|---|---|---|---|---|---|
| alcohol | /ˈæl.kə.hɒl/ | 7 | 7 | 3 | 4.52 | noun | 9.00 | 4.76 |
| alone | /əˈləʊn/ | 5 | 4 | 2 | 5.03 | adverb | 4.94 | 2.86 |
| always | /ˈɔːl.weɪz/ | 6 | 5 | 2 | 5.76 | adverb | 6.26 | 1.71 |
| ambulance | /ˈæm.bjə.ləns/ | 9 | 9 | 3 | 4.47 | noun | 6.16 | 4.81 |
| answer | /ˈɑːn.səʳ/ | 6 | 5 | 2 | 5.56 | noun | 5.43 | 2.89 |
| apple | /ˈæp.ᵊl/ | 5 | 3 | 2 | 4.58 | noun | 4.15 | 5 |
| arm | /ɑːm/ | 3 | 2 | 1 | 4.7 | noun | 3.26 | 4.96 |
| athletes | /ˈæθ.liːt/ | 8 | 6 | 2 | 4.43 | noun | n/a | n/a |
| bath | /bɑːθ/ | 4 | 3 | 1 | 4.65 | noun | 3.23 | 4.85 |
| beans | /biːn/ | 5 | 4 | 1 | 4.39 | noun | n/a | n/a |
| beautiful | /ˈbjuː.tɪ.fᵊl/ | 9 | 8 | 3 | 5.42 | adjective | 5.72 | 2.16 |
| bird | /bɜːd/ | 4 | 3 | 1 | 4.85 | noun | 3.52 | 5 |
| blouse | /blaʊz/ | 6 | 4 | 1 | 3.08 | noun | 6.65 | 4.96 |
| bottle | /ˈbɒt.ᵊl/ | 6 | 4 | 2 | 4.65 | noun | 3.56 | 4.91 |
| bread | /bred/ | 5 | 4 | 1 | 4.69 | noun | 3.58 | 4.92 |
| brush | /brʌʃ/ | 5 | 4 | 1 | 4.29 | noun | 3.78 | 4.54 |
| burger | /ˈbɜː.gəʳ/ | 6 | 5 | 2 | 3.90 | noun | 5.17 | 4.93 |
| card | /kɑːd/ | 4 | 3 | 1 | 4.89 | noun | 6.20 | 4.9 |
| careful | /ˈkeə.fᵊl/ | 7 | 5 | 3 | 4.84 | adjective | 5.09 | 1.86 |
| carrots | /ˈkær.ət/ | 7 | 6 | 2 | 4.18 | noun | n/a | n/a |
| case | /keɪs/ | 4 | 3 | 1 | 5.43 | noun | 6.74 | 3.93 |
| castle | /ˈkɑː.sᵊl/ | 6 | 4 | 2 | 4.68 | noun | 5.80 | 4.96 |
| cat | /kæt/ | 3 | 3 | 1 | 4.83 | noun | 3.68 | 4.86 |

| Word | British Pronunciation | Ch | Ph | Sy | Zipf | PoS | AoA | Con. |
|---|---|---|---|---|---|---|---|---|
| celery | /ˈsel.ᵊr.i/ | 6 | 6 | 3 | 3.66 | noun | 5.78 | 4.8 |
| character | /ˈkær.ək.təʳ/ | 9 | 8 | 3 | 4.93 | noun | 6.47 | 2.93 |
| charge | /tʃɑːdʒ/ | 6 | 3 | 1 | 4.94 | noun | 7.50 | 2.96 |
| cherry | /ˈtʃer.i/ | 6 | 4 | 2 | 4.23 | noun | 5.58 | 4.62 |
| chicken | /ˈtʃɪk.ɪn/ | 7 | 5 | 2 | 4.82 | noun | 3.26 | 4.8 |
| church | /tʃɜːtʃ/ | 6 | 3 | 1 | 5.02 | noun | 5.15 | 4.9 |
| clever | /ˈklev.əʳ/ | 6 | 6 | 2 | 4.71 | adjective | 7.50 | 1.79 |
| clock | /klɒk/ | 5 | 4 | 1 | 4.89 | noun | 4.42 | 5 |
| coat | /kəʊt/ | 4 | 3 | 1 | 4.41 | noun | 3.58 | 4.97 |
| cold | /kəʊld/ | 4 | 4 | 1 | 5.11 | adjective | 3.95 | 3.85 |
| confidence | /ˈkɒn.fɪ.dᵊns/ | 10 | 9 | 3 | 4.88 | noun | 9.28 | 2.17 |
| contact | /ˈkɒn.tækt/ | 7 | 7 | 2 | 4.72 | noun | 7.78 | 3.86 |
| create | /kriˈeɪt/ | 6 | 5 | 2 | 4.95 | verb | 7.10 | 2.62 |
| cross | /krɒs/ | 5 | 4 | 1 | 4.94 | noun | 4.74 | 4.44 |
| crowd | /kraʊd/ | 5 | 4 | 1 | 4.76 | noun | 7.14 | 4.52 |
| crown | /kraʊn/ | 5 | 4 | 1 | 4.53 | noun | 7.80 | 4.81 |
| cucumber | /ˈkjuː.kʌm.bəʳ/ | 8 | 9 | 3 | 3.77 | noun | 5.72 | 4.83 |
| cup | /kʌp/ | 3 | 3 | 1 | 5.09 | noun | 3.57 | 5 |
| danger | /ˈdeɪn.dʒəʳ/ | 6 | 6 | 2 | 4.75 | noun | 4.61 | 2.68 |
| dark | /dɑːk/ | 4 | 3 | 1 | 4.94 | adjective | 3.74 | 4.29 |
| date | /deɪt/ | 4 | 3 | 1 | 4.94 | noun | 5.84 | 3.9 |
| deep | /diːp/ | 4 | 3 | 1 | 4.98 | adjective | 5.47 | 3.38 |
| dentist | /ˈden.tɪst/ | 7 | 7 | 2 | 3.76 | noun | 5.22 | 4.93 |

| Word | British Pronunciation | Ch | Ph | Sy | Zipf | PoS | AoA | Con. |
|---|---|---|---|---|---|---|---|---|
| dinner | /ˈdɪn.əʳ/ | 6 | 5 | 2 | 4.77 | noun | 3.99 | 4.5 |
| direction | /daɪˈrek.ʃᵊn/ | 9 | 7 | 3 | 4.69 | noun | 6.68 | 2.79 |
| disease | /dɪˈziːz/ | 7 | 5 | 2 | 4.49 | noun | 7.55 | 3.45 |
| distance | /ˈdɪs.tᵊns/ | 8 | 7 | 2 | 4.64 | noun | 6.81 | 3.17 |
| doctor | /ˈdɒk.təʳ/ | 6 | 6 | 2 | 5.02 | noun | 4.60 | 4.69 |
| door | /dɔːʳ/ | 4 | 3 | 1 | 5.26 | noun | 3.05 | 4.81 |
| down | /daʊn/ | 4 | 3 | 1 | 6.09 | preposition | 4.93 | 3.52 |
| downstairs | /ˌdaʊnˈsteəz/ | 10 | 7 | 2 | 4.53 | noun | 3.91 | 4.43 |
| drive | /draɪv/ | 5 | 4 | 1 | 5 | verb | 5.26 | 3.86 |
| drop | /drɒp/ | 4 | 4 | 1 | 4.9 | verb | 3.26 | 4.21 |
| early | /ˈɜː.li/ | 5 | 3 | 2 | 5.11 | adverb | 5.31 | 2.25 |
| easy | /ˈiː.zi/ | 4 | 3 | 2 | 4.42 | adjective | 5.43 | 2.07 |
| engine | /ˈen.dʒɪn/ | 6 | 5 | 2 | 4.5 | noun | 6.28 | 4.86 |
| enjoy | /ɪnˈdʒɔɪ/ | 5 | 4 | 2 | 5.05 | verb | 5.75 | 2.29 |
| enough | /ɪˈnʌf/ | 6 | 4 | 2 | 5.65 | adverb | 6.96 | 1.33 |
| evening | /ˈiːv.nɪŋ/ | 7 | 5 | 2 | 5.12 | noun | 5.84 | 3.26 |
| evidence | /ˈev.ɪ.dᵊns/ | 8 | 7 | 3 | 5.06 | noun | 10.58 | 3.65 |
| fashion | /ˈfæʃ.ᵊn/ | 7 | 4 | 2 | 4.63 | noun | 8.20 | 3.43 |
| feet | /fiːt/ | 4 | 3 | 1 | 5.08 | noun | n/a | n/a |
| female | /ˈfiː.meɪl/ | 6 | 5 | 2 | 4.67 | adjective | 5.89 | 4.57 |
| finance | /ˈfaɪ.næns/ | 7 | 6 | 2 | 4.41 | noun | 12.52 | 3.36 |
| finish | /ˈfɪn.ɪʃ/ | 6 | 5 | 2 | 5.06 | noun | n/a | n/a |
| flag | /flæg/ | 4 | 4 | 1 | 4.42 | noun | n/a | n/a |
| flavour | /ˈfleɪ.vəʳ/ | 7 | 6 | 2 | 4.73 | noun | n/a | n/a |
| fleece | /fliːs/ | 6 | 4 | 1 | 3.37 | noun | 10.2 | 4.75 |
| football | /ˈfʊt.bɔːl/ | 8 | 6 | 2 | 5.12 | noun | 4.84 | 4.73 |
| forest | /ˈfɒr.ɪst/ | 6 | 6 | 2 | 4.67 | noun | 6.28 | 4.76 |
| fork | /fɔːk/ | 4 | 3 | 1 | 3.97 | noun | 3.63 | 4.9 |
| frame | /freɪm/ | 5 | 4 | 1 | 4.76 | noun | 7.67 | 4.3 |
| freedom | /ˈfriː.dəm/ | 7 | 6 | 2 | 4.64 | noun | 7.05 | 2.34 |
| full | /fʊl/ | 4 | 3 | 1 | 5.4 | adjective | 4.24 | 3.59 |
| funny | /ˈfʌn.i/ | 5 | 4 | 2 | 5.06 | adjective | 5.47 | 2.5 |
| future | /ˈfjuː.tʃəʳ/ | 6 | 6 | 2 | 5.3 | noun | 7.16 | 1.86 |
| gift | /gɪft/ | 4 | 4 | 1 | 4.44 | noun | 5.05 | 4.56 |
| girl | /gɜːl/ | 4 | 3 | 1 | 5.29 | noun | 4.00 | 4.85 |

| Word | British Pronunciation | Ch | Ph | Sy | Zipf | PoS | AoA | Con. |
|---|---|---|---|---|---|---|---|---|
| glasses | /ˈglɑː.sɪz/ | 7 | 6 | 2 | 4.37 | Noun | 4.35 | 4.9 |
| golf | /gɒlf/ | 4 | 4 | 1 | 4.52 | Noun | 7.16 | 4.52 |
| ground | /graʊnd/ | 6 | 5 | 1 | 5.19 | Noun | 4.89 | 4.77 |
| hair | /heəʳ/ | 4 | 3 | 1 | 5.03 | Noun | 3.17 | 4.97 |
| handbag | /ˈhænd.bæg/ | 7 | 7 | 2 | 3.73 | Noun | 7.95 | 4.93 |
| he | /hiː/ | 2 | 2 | 1 | 6.82 | pronoun | 3.81 | 3.93 |
| heart | /hɑːt/ | 5 | 3 | 1 | 5.3 | noun | 5.17 | 4.52 |
| heavy | /ˈhev.i/ | 5 | 4 | 2 | 4.9 | adjective | 4.05 | 3.37 |
| help | /help/ | 4 | 4 | 1 | 5.75 | verb | 3.65 | 2.56 |
| hide | /haɪd/ | 4 | 3 | 1 | 4.66 | verb | 4.47 | 3.21 |
| hole | /həʊl/ | 4 | 3 | 1 | 4.76 | noun | 5.05 | 4.81 |
| hot | /hɒt/ | 3 | 3 | 1 | 5.14 | adjective | 3.37 | 4.31 |
| hour | /aʊəʳ/ | 4 | 3 | 2 | 5.17 | noun | 5.85 | 3.1 |
| ill | /ɪl/ | 3 | 2 | 1 | 4.59 | adjective | 5.60 | 3.21 |
| important | /ɪmˈpɔː.tᵊnt/ | 9 | 7 | 3 | 5.57 | adjective | 5.79 | 2.14 |
| improve | /ɪmˈpruːv/ | 7 | 6 | 2 | 4.61 | verb | 7.47 | 1.82 |
| interest | /ˈɪn.trəst/ | 8 | 7 | 2 | 5.05 | noun | 7.37 | 1.8 |
| interview | /ˈɪn.tə.vjuː/ | 9 | 7 | 3 | 4.59 | noun | 10.53 | 3.36 |
| island | /ˈaɪ.lənd/ | 6 | 5 | 2 | 4.96 | noun | 7.41 | 4.96 |
| joke | /dʒəʊk/ | 4 | 3 | 1 | 4.59 | noun | 5.20 | 2.9 |
| judge | /dʒʌdʒ/ | 5 | 3 | 1 | 4.69 | noun | 8.85 | 3.75 |
| juice | /dʒuːs/ | 5 | 3 | 1 | 4.44 | noun | 4.40 | 4.89 |
| kettle | /ˈket.ᵊl/ | 6 | 4 | 2 | 4.02 | noun | 8.06 | 4.75 |
| kiss | /kɪs/ | 4 | 3 | 1 | 4.65 | verb | 3.61 | 4.48 |
| knife | /naɪf/ | 5 | 3 | 1 | 4.49 | noun | 4.15 | 4.9 |
| landscape | /ˈlænd.skeɪp/ | 9 | 8 | 2 | 4.62 | noun | 9.89 | 4.34 |
| lasagna | /ləˈzæn.jə/ | 7 | 7 | 3 | 1.74 | noun | 5 | 4.89 |
| late | /leɪt/ | 4 | 3 | 1 | 5.21 | adjective | 5.28 | 2.6 |
| lawnmower | /ˈlɔːnˌməʊ.əʳ/ | 9 | 6 | 3 | 3.11 | noun | 6.11 | 4.97 |
| leader | /ˈliː.dəʳ/ | 6 | 5 | 2 | 5.02 | noun | 6.90 | 3.89 |
| learn | /lɜːn/ | 5 | 3 | 1 | 5.01 | verb | 4.44 | 2.2 |
| line | /laɪn/ | 4 | 3 | 1 | 5.45 | noun | 4.85 | 4.5 |
| little | /ˈlɪt.ᵊl/ | 6 | 4 | 2 | 6.03 | adjective | 3.95 | 3.67 |
| love | /lʌv/ | 4 | 3 | 1 | 5.88 | verb | 5.17 | 2.07 |
| mad | /mæd/ | 3 | 3 | 1 | 4.73 | adjective | 3.55 | 2.76 |

| Word | British Pronunciation | Ch | Ph | Sy | Zipf | PoS | AoA | Con. |
|---|---|---|---|---|---|---|---|---|
| meal | /mɪəl/ | 4 | 3 | 1 | 4.53 | noun | 4.74 | 4.62 |
| message | /ˈmes.ɪdʒ/ | 7 | 5 | 2 | 4.94 | noun | 6.32 | 3.97 |
| mile | /maɪl/ | 4 | 3 | 1 | 4.62 | noun | 7.65 | 3.63 |
| mind | /maɪnd/ | 4 | 4 | 1 | 5.47 | noun | 5.37 | 2.5 |
| mirror | /ˈmɪr.əʳ/ | 6 | 5 | 2 | 4.45 | noun | 4.89 | 4.97 |
| mobile | /ˈməʊ.baɪl/ | 6 | 5 | 2 | 4.57 | adjective | 8.72 | 2.93 |
| modern | /ˈmɒd.ᵊn/ | 6 | 4 | 2 | 5.02 | adjective | 8.58 | 2.31 |
| motorbike | /ˈməʊ.tə.baɪk/ | 9 | 7 | 3 | 3.71 | noun | 7.95 | 4.79 |
| name | /neɪm/ | 4 | 3 | 1 | 5.61 | noun | 3.68 | 3.5 |
| neighbours | /ˈneɪ.bəʳ/ | 10 | 5 | 2 | 4.48 | noun | n/a | n/a |
| nervous | /ˈnɜː.vəs/ | 7 | 5 | 2 | 4.87 | adjective | 6.79 | 1.85 |
| news | /njuːz/ | 4 | 4 | 2 | 5.41 | noun | 6.16 | 3.41 |
| noise | /nɔɪz/ | 5 | 3 | 1 | 5.08 | noun | 4.50 | 3.52 |
| nurse | /nɜːs/ | 5 | 3 | 1 | 4.5 | noun | 5.84 | 4.39 |
| old | /əʊld/ | 3 | 3 | 1 | 5.79 | adjective | 3.72 | 2.72 |
| orange | /ˈɒr.ɪndʒ/ | 6 | 5 | 2 | 4.64 | noun | 3.26 | 4.66 |
| pack | /pæk/ | 4 | 3 | 1 | 4.62 | noun | 6.55 | 3.83 |
| patient | /ˈpeɪ.ʃᵊnt/ | 7 | 5 | 2 | 4.72 | noun | 7.05 | 2.5 |
| pay | /peɪ/ | 3 | 2 | 1 | 5.4 | verb | 5.50 | 3.54 |
| personal | /ˈpɜː.sᵊn.ᵊl/ | 8 | 6 | 3 | 4.89 | adjective | 8.05 | 2.21 |
| picture | /ˈpɪk.tʃəʳ/ | 7 | 6 | 2 | 5.13 | noun | 4.05 | 4.52 |
| pizza | /ˈpiːt.sə/ | 5 | 5 | 2 | 4.25 | noun | 4.67 | 5 |
| police | /pəˈliːs/ | 6 | 5 | 2 | 5.5 | noun | 5.05 | 4.79 |
| poor | /pɔːʳ/ | 4 | 3 | 1 | 5.05 | adjective | 6.11 | 2.7 |
| positive | /ˈpɒz.ə.tɪv/ | 8 | 7 | 3 | 4.8 | adjective | 8.11 | 2.44 |
| pregnant | /ˈpreg.nənt/ | 8 | 8 | 2 | 4.43 | adjective | 6.76 | 4.59 |
| programme | /ˈprəʊ.græm/ | 9 | 7 | 2 | 5.2 | noun | n/a | n/a |
| progress | /ˈprəʊ.gres/ | 8 | 7 | 2 | 4.68 | noun | 9.00 | 2.12 |
| protect | /prəˈtekt/ | 7 | 7 | 2 | 4.81 | verb | 6.79 | 2.86 |
| quality | /ˈkwɒl.ə.ti/ | 7 | 7 | 3 | 5.01 | noun | 8.78 | 2.18 |
| quick | /kwɪk/ | 5 | 4 | 1 | 5.13 | adjective | 5.89 | 2.89 |
| race | /reɪs/ | 4 | 3 | 1 | 5.11 | noun | 6.00 | 3.59 |
| radio | /ˈreɪ.di.əʊ/ | 5 | 5 | 3 | 4.82 | noun | 5.17 | 4.74 |
| return | /rɪˈtɜːn/ | 6 | 5 | 2 | 4.91 | noun | 5.61 | 2.97 |
| ring | /rɪŋ/ | 4 | 3 | 1 | 4.92 | noun | 4.53 | 4.81 |

| Word | British Pronunciation | Ch | Ph | Sy | Zipf | PoS | AoA | Con. |
|---|---|---|---|---|---|---|---|---|
| robin | /ˈrɒb.ɪn/ | 5 | 5 | | 4.42 | noun | 6.69 | 4.61 |
| rose | /rəʊz/ | 4 | 3 | | 4.64 | noun | 6.11 | 4.9 |
| rough | /rʌf/ | 5 | 3 | 1 | 4.46 | adjective | 6.21 | 3.83 |
| run | /rʌn/ | 3 | 3 | 1 | 5.49 | verb | 4.47 | 4.31 |
| salad | /ˈsæl.əd/ | 5 | 5 | | 4.38 | noun | 5.61 | 4.97 |
| same | /seɪm/ | 4 | 3 | 1 | 5.65 | adjective | 5.62 | 2.64 |
| sand | /sænd/ | 4 | 4 | 1 | 4.51 | noun | 4.63 | 5 |
| scared | /skeəd/ | 6 | 4 | 1 | 4.81 | verb | 3.79 | 2.5 |
| school | /skuːl/ | 6 | 4 | 1 | 5.45 | noun | 3.89 | 4.79 |
| secret | /ˈsiː.krət/ | 6 | 6 | 2 | 4.93 | adjective | 5.39 | 2.19 |
| shape | /ʃeɪp/ | 5 | 3 | 1 | 4.9 | noun | 4.47 | 3.14 |
| sharp | /ʃɑːp/ | 5 | 3 | 1 | 4.55 | adjective | 6.11 | 3.86 |
| she | /ʃiː/ | 3 | 2 | 1 | 6.42 | pronoun | 3.57 | 3.36 |
| shops | /ˈbɑː.bə.ʃɒp/ | 5 | 4 | 1 | 4.6 | noun | n/a | n/a |
| short | /ʃɔːt/ | 5 | 3 | 1 | 5.13 | adjective | 4.32 | 3.61 |
| shout | /ʃaʊt/ | 5 | 3 | 1 | 4.43 | verb | 4.72 | 4.17 |
| sight | /saɪt/ | 5 | 3 | 1 | 4.64 | noun | 5.44 | 3.21 |
| sitting | /ˈsɪt.ɪŋ/ | 7 | 5 | 2 | 5.01 | verb | n/a | 3.76 |
| sky | /skaɪ/ | 3 | 3 | 1 | 4.79 | noun | 4.17 | 4.45 |
| snow | /snəʊ/ | 4 | 3 | 1 | 4.79 | noun | 4.11 | 4.85 |
| song | /sɒŋ/ | 4 | 3 | 1 | 5.1 | noun | 4.26 | 4.46 |
| sound | /saʊnd/ | 5 | 4 | 1 | 5.16 | noun | 3.72 | 3.7 |
| soup | /suːp/ | 4 | 3 | 1 | 4.41 | noun | 5.37 | 4.72 |
| special | /ˈspeʃ.ᵊl/ | 7 | 5 | 2 | 5.32 | adjective | 5.00 | 1.76 |
| speech | /spiːtʃ/ | 6 | 4 | 1 | 4.72 | noun | 6.22 | 3.37 |
| speed | /spiːd/ | 5 | 4 | 1 | 4.91 | noun | 5.11 | 3.62 |
| spring | /sprɪŋ/ | 6 | 5 | 1 | 4.69 | noun | 5.50 | 3.89 |
| staff | /stɑːf/ | 5 | 4 | 1 | 4.95 | noun | 10.00 | 4.36 |
| star | /stɑːʳ/ | 4 | 4 | 1 | 5.04 | noun | 3.89 | 4.69 |
| start | /stɑːt/ | 5 | 4 | 1 | 5.77 | verb | 4.37 | 2.71 |
| station | /ˈsteɪ.ʃᵊn/ | 7 | 5 | 2 | 4.82 | noun | 7.11 | 4.32 |
| steak | /steɪk/ | 5 | 4 | | 4.05 | noun | 6.63 | 4.96 |
| street | /striːt/ | 6 | 5 | 1 | 5.2 | noun | 4.58 | 4.75 |
| student | /ˈstjuː.dᵊnt/ | 7 | 7 | 2 | 4.61 | noun | 5.94 | 4.92 |
| style | /staɪl/ | 5 | 4 | 1 | 4.98 | noun | 7.58 | 2.67 |
| success | /səkˈses/ | 7 | 6 | 2 | 4.92 | noun | 9.25 | 2.21 |

| Word | British Pronunciation | Ch | Ph | Sy | Zipf | PoS | AoA | Con. |
|---|---|---|---|---|---|---|---|---|
| summer | /ˈsʌm.əʳ/ | 6 | 5 | 2 | 5 | noun | 4.33 | 3.64 |
| sun | /sʌn/ | 3 | 3 | 1 | 5.01 | noun | 3.40 | 4.83 |
| table | /ˈteɪ.bᵊl/ | 5 | 4 | 2 | 5.1 | noun | 4.39 | 4.9 |
| take | /teɪk/ | 4 | 3 | 1 | 6.08 | verb | 4.37 | 3.06 |
| talking | /tɔːk/ | 7 | 5 | 2 | 5.47 | verb | n/a | 3.72 |
| tax | /tæks/ | 3 | 4 | 1 | 5.18 | noun | 8.94 | 3.89 |
| taxi | /ˈtæk.si/ | 4 | 5 |  | 4.30 | noun | 7.58 | 4.93 |
| teacher | /ˈtiː.tʃəʳ/ | 7 | 5 | 2 | 4.68 | noun | 4.55 | 4.52 |
| teeth | /tiːθ/ | 5 | 3 | 1 | 4.7 | noun | n/a | 4.96 |
| think | /θɪŋk/ | 5 | 4 | 1 | 6.51 | verb | 4.75 | 2.41 |
| toaster | /ˈtəʊ.stəʳ/ | 7 | 6 |  | 3.29 | noun | 6.72 | 4.9 |
| toilet | /ˈtɔɪ.lət/ | 6 | 5 | 2 | 4.54 | noun | 3.54 | 4.97 |
| train | /treɪn/ | 5 | 4 | 1 | 4.98 | noun | 4.00 | 4.79 |
| trainers | /ˈtreɪ.nəʳ/ | 8 | 6 |  | 3.88 | noun | n/a | n/a |
| transport | /ˈtræn.spɔːt/ | 9 | 8 | 2 | 4.62 | noun | 10.38 | 3.5 |
| travel | /ˈtræv.ᵊl/ | 6 | 5 | 2 | 4.82 | verb | 5.90 | 3.67 |
| tree | /triː/ | 4 | 3 | 1 | 4.95 | noun | 3.57 | 5 |

| Word | British Pronunciation | Ch | Ph | Sy | Zipf | PoS | AoA | Con. |
|---|---|---|---|---|---|---|---|---|
| trouble | /ˈtrʌb.ᵊl/ | 7 | 5 | 2 | 5.08 | noun | 4.56 | 2.25 |
| up | /ʌp/ | 2 | 2 | 1 | 6.53 | preposition | 2.92 | 3.83 |
| upset | /ʌpˈset/ | 5 | 5 | 2 | 4.69 | adjective | 5.26 | 2.41 |
| upstairs | /ʌpˈsteəz/ | 8 | 6 | 2 | 4.75 | noun | 3.86 | 4.33 |
| waiting | /weɪt/ | 7 | 5 | 2 | 5.13 | verb | n/a | 2.7 |
| war | /wɔːʳ/ | 3 | 3 | 1 | 5.34 | noun | 7.67 | 3.63 |
| watch | /wɒtʃ/ | 5 | 3 | 1 | 5.3 | verb | 4.33 | 4.61 |
| water | /ˈwɔː.təʳ/ | 5 | 5 | 2 | 5.53 | noun | 2.37 | 5 |
| week | /wiːk/ | 4 | 3 | 1 | 5.66 | noun | 5.11 | 3.48 |
| wheel | /wiːl/ | 5 | 3 | 1 | 4.51 | noun | 4.40 | 4.86 |
| white | /waɪt/ | 5 | 3 | 1 | 5.25 | adjective | 4.06 | 3.89 |
| winning | /ˈwɪn.ɪŋ/ | 7 | 5 | 2 | 4.97 | verb | n/a | 2.82 |
| winter | /ˈwɪn.təʳ/ | 6 | 6 | 2 | 4.83 | noun | 4.38 | 3.84 |
| wonder | /ˈwʌn.dəʳ/ | 6 | 6 | 2 | 5.06 | verb | 7.63 | 1.8 |
| working | /ˈwɜː.kɪŋ/ | 7 | 5 | 2 | 5.51 | verb | n/a | 3.48 |
| write | /raɪt/ | 5 | 3 | 1 | 4.89 | verb | 4.89 | 4.22 |